\documentclass[10pt]{article}
\usepackage[preprint]{tmlr}

\usepackage{amsmath,amsfonts,amssymb,amsthm}
\usepackage{graphicx}
\usepackage{booktabs}
\usepackage{array}
\usepackage{float}
\usepackage[section]{placeins}
\usepackage{xcolor}
\usepackage{rotating}
\usepackage{algorithm}
\usepackage[shortlabels]{enumitem}
\usepackage[noend]{algpseudocode}
\usepackage{tikz}
\usetikzlibrary{
    arrows.meta,
    positioning,
    decorations.pathreplacing,
    fit,
    calc,
    shapes.geometric,
    backgrounds
}
\usepackage{hyperref}
\usepackage{url}
\usepackage{multirow, makecell}
\hypersetup{hidelinks}

\newcommand{\dif}{\ensuremath{\mathrm{d}}}

\newcommand{\T}{\ensuremath{\mathrm{\scriptscriptstyle T}}}
\usepackage{bm}

\newcommand{\Thetab}{\ensuremath{\bm\Theta}}

\newcommand{\Pib}{\ensuremath{\bm\Pi}}
\newcommand{\Sigmab}{\ensuremath{\bm\Sigma}}

\newcommand{\etab}{\ensuremath{\bm\eta}}
\newcommand{\thetab}{\ensuremath{\bm\theta}}

\newcommand{\xib}{\ensuremath{\bm\xi}}
\newcommand{\pib}{\ensuremath{\bm\pi}}

\newcommand{\sigmab}{\ensuremath{\bm\sigma}}
\newcommand{\taub}{\ensuremath{\bm\tau}}

\newcommand{\psib}{\ensuremath{\bm\psi}}

\newcommand{\varepsilonb}{\ensuremath{\bm\varepsilon}}

\newcommand{\varphib}{\ensuremath{\bm\varphi}}

\newcommand{\bmb}{\ensuremath{\bm b}}
\newcommand{\bmc}{\ensuremath{\bm c}}

\newcommand{\bme}{\ensuremath{\bm e}}

\newcommand{\bmg}{\ensuremath{\bm g}}

\newcommand{\bmp}{\ensuremath{\bm p}}
\newcommand{\bmq}{\ensuremath{\bm q}}
\newcommand{\bmr}{\ensuremath{\bm r}}

\newcommand{\bmx}{\ensuremath{\bm x}}

\newcommand{\bmB}{\ensuremath{\bm B}}

\newcommand{\bmH}{\ensuremath{\bm H}}

\newcommand{\bmK}{\ensuremath{\bm K}}

\newcommand{\bmM}{\ensuremath{\bm M}}

\newcommand{\bmQ}{\ensuremath{\bm Q}}

\newcommand{\bmV}{\ensuremath{\bm V}}
\newcommand{\bmW}{\ensuremath{\bm W}}
\newcommand{\bmX}{\ensuremath{\bm X}}
\newcommand{\bmY}{\ensuremath{\bm Y}}
\newcommand{\bmZ}{\ensuremath{\bm Z}}
\usepackage{amsmath}

\DeclareMathOperator*{\argmax}{arg\,max}

\usepackage{mathrsfs}

\usepackage{color}

\definecolor{myorange}{rgb}{0.87843, 0.70588, 0.18824}

\definecolor{primaryred}{RGB}{213, 0, 50}
\definecolor{softred}{RGB}{241,169,160}
\definecolor{primarynavy}{RGB}{0, 79, 113}
\definecolor{softnavy}{RGB}{142,162,198}
\definecolor{primaryorange}{RGB}{230,119,0}
\definecolor{softorange}{RGB}{250,192,144}

\renewcommand{\arraybackslash}{\let\\\tabularnewline\scriptsize}

\newtheorem{theorem}{Theorem}[section]
\newtheorem{lemma}[theorem]{Lemma}

\title{Generative Modeling for Cascading Extremes via\\
Autoregressive Transformers}

\author{
    \name \hspace{-0.3cm} Miguel de Carvalho \email miguel.decarvalho@ed.ac.uk \\
    \addr School of Mathematics and Maxwell Institute for Mathematical Sciences\\
    University of Edinburgh \\
    CIDMA--Department of Mathematics \\
    Universidade de Aveiro \vspace{-0.3cm}
    \AND
    \name {Myung Won} Lee \email johnny.myungwon.lee@ed.ac.uk \\
    \addr School of Mathematics and Maxwell Institute for Mathematical Sciences\\
    University of Edinburgh \vspace{-0.3cm}
    \AND
    \name Jordan Richards \email jordan.richards@ed.ac.uk \\
    \addr School of Mathematics and Maxwell Institute for Mathematical Sciences\\
    University of Edinburgh \vspace{-0.3cm}
    \AND
    \name Sotirios Sabanis \email s.sabanis@ed.ac.uk \\
    \addr School of Mathematics and Maxwell Institute for Mathematical Sciences\\
    University of Edinburgh \\
    National Technical University of Athens \\
    Athena/Archimedes Research Centre
}

\def\month{MM}
\def\year{YYYY}
\def\openreview{\url{https://openreview.net/forum?id=XXXX}}

\begin{document}
\maketitle

\begin{abstract}
Extreme events often occur sequentially, from earthquake--tsunami cascades to successive extreme losses in financial markets. We introduce an extremal token representation that encodes multivariate observations as a categorical sequence. A quiet token records the absence of threshold exceedances; otherwise, the token identifies the exceeding component with the largest marginal probability-scale score. The representation is invariant to componentwise strictly increasing transformations. It preserves the identity and temporal ordering of dominant exceedances but discards their magnitudes and the identities of nondominant simultaneous exceedances. A decoder-only transformer estimates the history-dependent next-token kernel on this representation, and recursive sampling generates future token sequences from the induced joint predictive law. We amortize our generative model by pretraining across simulated process families with varying parameters and dependence persistence. Next-token predictions are evaluated against exact conditional probabilities for held-out categorical processes and particle filter references for heavy-tailed vector autoregressive processes. Applied to ten-second returns for five semiconductor stocks, the pretrained model supports zero-shot transfer, while a few epochs of adaptation improve agreement between observed and generated token frequencies and transition patterns. These results support task-specific tokenization and simulation-based pretraining for modeling which dimensions dominate multivariate extremes.
\end{abstract}
\setcounter{table}{1}
\section{Introduction}\label{intro}

Extreme events often unfold sequentially. As recently emphasized by \citet{decarvalho2026kolmogorov}, the order in which such events occur matters in many applications, and modeling such sequences of extremes poses nontrivial methodological challenges. Building on their work, we develop this perspective through an analogy with text: Just as a large language model generates a continuation conditional on preceding words, we seek to generate sequences of extremes conditional on their recent history.

While generative modeling of extremes is an increasingly important task in modern machine learning and extreme value theory \citep[e.g.,][]{allouche2022ev, de2025generative}, we focus on how extreme activity evolves across components over time. We study the predictive law of what we will refer to as \emph{extremal token sequences}. Given recent extreme activity, we predict the next quiet or dominant-exceedance token and assess plausible subsequent token paths.

We encode the multivariate observations as extremal tokens. Each observation receives a quiet token when no threshold is exceeded; otherwise, it records the exceeding component, selecting the one with the greatest marginal probability-scale score when multiple components exceed. For a $d$-variate series, the vocabulary contains $d+1$ tokens. The representation retains quiet observations and the ordered labels of dominant exceedances, while discarding their magnitudes and nondominant simultaneous exceedances. We predict and generate these token sequences rather than the original continuous series. An autoregressive model supports both prediction and generation in this representation. It specifies next-token conditional probabilities; their product defines the joint law from which future paths are sampled recursively. We use a decoder-only transformer to learn this next-token kernel. To reduce dependence on tail observations from any one series, we pretrain a shared model on simulated processes spanning different parameters and orders. This amortizes training across prediction tasks, enabling direct use on new series or further adaptation.

\noindent\textbf{Main contributions.} 
\noindent Our key contributions are as follows: 
\begin{enumerate}[nolistsep]
    \item We pioneer the development of generative models for cascading extremes by 
    introducing a $(d+1)$-token representation for extremal sequences, that is invariant to componentwise strictly increasing transformations of the raw multivariate time series. \vspace{0.1cm}
    \item We characterize the high-threshold limit of the next-token kernel using extreme value theory and bound fixed-horizon recursive-generation error in terms of one-step approximation error for finite-context kernels.\vspace{0.1cm}
    \item We develop an amortized next-token learning scheme on simulated data and evaluate it on independently sampled processes from the training families, using exact conditional probabilities for categorical processes and sequential Monte Carlo references for tokenized heavy-tailed vector autoregressions.\vspace{0.1cm}
    \item We investigate zero-shot transfer and adaptation using ten-second negative returns for five semiconductor stocks, assessing agreement between observed and generated token frequencies and transition structure.
\end{enumerate}

\noindent\textbf{Related prior work.}
Extreme value theory provides the probabilistic framework for studying univariate and multivariate extremes. Our theoretical development draws on this framework to characterize the limiting conditional probabilities of dominant-exceedance tokens.

The mathematical foundations of artificial intelligence (AI) can be traced to developments in logic and computability in the 1930s \citep{ertel2025introduction}, roughly the same period in which extreme value theory (EVT) emerged through the pioneering work of Fisher and Tippett \citep{fisher1928}. Despite these parallel beginnings, it has taken nearly a century for the two fields to begin meaningfully intersecting. The special issue editorial by \citet{de2026bridging} surveys the emerging interface between heavy tails and artificial intelligence. \citet{clemencon2026weak} review key results connecting multivariate extreme value theory and statistical learning theory. 

Deep generative models, including generative adversarial networks (GANs) \citep{goodfellow2014generative}, normalizing flows \citep{papamakarios2021normalizing}, and diffusion models \citep{ho2020denoising} have been supported by theoretically in approximation and convergence \citep{liang2021well, teshima2020coupling, bruno2025wasserstein}. However, standard implementations 
can encounter difficulties in accurately reproducing heavy tails and extremal dependence. These challenges have motivated generative methods tailored to extremes, including GAN-based approaches \citep{bhatia2021exgan,allouche2022ev,allouche2026exceedgan,lhaut2026simulation} and normalizing flows \citep{de2025generative,hu2025gpdflow}. In parallel, diffusion models have been explored for extreme-weather forecasting \citep{price2023gencast}  and the generation of financial time series and spatial fields exhibiting heavy tails \citep{takahashi2025generation, walchessen2025neural}. Complementing these developments, \citet{wessel2026generative} compare the generative models for multivariate angular simulation.

A key distinction in our setting, with respect to that of previous works, lies in the generated object. Our model produces categorical paths of what we here define as quiet and dominant-exceedance tokens, rather than continuous observations. The predictive target is the history-dependent distribution of the next token in this representation. To achieve this, we build upon an autoregressive transformer architecture \citep{vaswani2017attention,radford2018improving,bishop2024deep} and on related work in amortized learning and neural inference \citep{ruoss2024amortized,richards2024neural,zammit2025neural,lee2026spatial}. Although the model describes how predictions depend on preceding extreme activity, these conditional distributions do not identify causal effects \citep{pearl2009causality,gnecco2021causal}.

\noindent\textbf{Outline.}
Section~\ref{sec:trans} introduces the extremal-token representation and the high-threshold limit of its next-token kernel. It also details the transformer model and amortized training objective, and establishes fixed horizon stability of recursive generation. Section~\ref{sec:application} examines transferability of the amortized generator to semiconductor stock returns, while Section~\ref{sec:conclusion} discusses limitations and extensions. \appendixname~\ref{appdx:theory} provides proofs and technical details, \appendixname~\ref{appdx:numexp} presents Monte Carlo simulation experiments, and the remaining appendices contain implementation details and diagnostics.

\section{Autoregressive Transformer for Cascading Extremes}\label{sec:trans}
\subsection{Probabilistic Setup and Extremal Tokens}\label{subsec:tech}
\subsubsection*{Framework}
Let $\{\bmY_t\}_{t=1}^n$ be a $d$-variate process, such as negative returns on a $d$-asset portfolio, with $\bmY_t=(Y_{t,1},\ldots,Y_{t,d})^\T$ and continuous marginal distribution functions $F_j(y)=\Pr(Y_{t,j}\leq y), j=1,\dots,d$. Choose thresholds $u_j=F_j^{-1}(q)$ at a common high probability level $q\in(0,1)$ (e.g., $q=0.98$), and define the extremal indicators $I_{t,j}=I(Y_{t,j}>u_j)$ and active sets $\mathcal{E}_t=\{j:I_{t,j}=1\}$ for each time $t\in\{1,\dots,n\}$.

We map the observations to a sequence of \emph{extremal tokens}, $\{A_t\}_{t=1}^n$, over the \emph{extremal vocabulary} $\mathcal{V}=\{0,1,\ldots,d\}$, where
\begin{equation}\label{eq:token}
    A_t= A_t(q) = 
    \begin{cases}
        0, & \mathcal{E}_t=\emptyset,\\
        \argmax\limits_{j\in\mathcal{E}_t} F_j\!\left(Y_{t,j}\right), & \mathcal{E}_t\neq\emptyset,
    \end{cases}
\end{equation}
Token $0$ is the \textit{quiet token}, indicating that there are no exceedances for any component $j$; otherwise, token $A_t \in \mathcal{V}$ identifies the dominant exceeding component. When only one component exceeds its threshold---as occurs at approximately $75\%$ of times with an exceedance in our data application---the token simply identifies the component experiencing the extreme. Thresholding determines which components are active, while the scores $F_j(Y_{t,j})$ rank them by marginal extremeness on a common uniform scale. With marginals and thresholds transformed consistently, these scores, and hence $A_t$, are invariant to componentwise strictly increasing transformations. \appendixname~\ref{appdx:proofs} gives the invariance argument.

\begin{figure}[!htbp]
\centering
\resizebox{\textwidth}{!}{%
\begin{tikzpicture}[
    >=Latex,
    font=\small,
    node distance=1.2cm and 1.6cm,
    box/.style={
        draw,
        rounded corners=3pt,
        thick,
        align=center,
        inner sep=6pt,
        fill=gray!6
    },
    stage/.style={
        draw,
        rounded corners=4pt,
        thick,
        align=center,
        inner sep=8pt,
        fill=teal!15
    },
    token/.style={
        draw,
        rounded corners=2pt,
        minimum width=0.65cm,
        minimum height=0.55cm,
        align=center,
        inner sep=1pt,
        thick
    },
    quiettoken/.style={token, fill=gray!18},
    activetoken/.style={token, fill=red!50},
    subbox/.style={
        draw,
        rounded corners=2pt,
        thick,
        align=center,
        inner sep=4pt,
        fill=white
    },
    flow/.style={->, thick},
    lab/.style={anchor=west, font=\scriptsize}
]


\node[
    stage,
    minimum width=6.0cm,
    minimum height=4.4cm
] (process) {};

\node[
    font=\bfseries\normalsize,
    anchor=north
] at ([yshift=-0.15cm]process.north)
{Multivariate stochastic process};

\coordinate (pSW)
    at ([xshift=-2.3cm,yshift=-1.5cm]process.center);

\draw[thick,->]
    (pSW) -- ++(0,2.8);

\draw[thick,->]
    (pSW) -- ++(4.75,0);

\begin{scope}[shift={(pSW)}]


\draw[
    red,
    dashed,
    line width=2.5pt
]
    (0,1.75) -- (4.55,1.75);


\draw[
    blue!70!black,
    thick,
    smooth
]
plot coordinates {
    (0.15,0.55)
    (0.60,1.20)
    (1.00,0.35)
    (1.40,1.05)
    (1.80,2.10)
    (2.15,0.85)
    (2.55,2.25)
    (3.00,0.70)
    (3.45,1.35)
    (3.85,1.00)
    (4.15,1.55)
    (4.45,0.90)
};


\draw[
    green!60!black,
    thick,
    smooth
]
plot coordinates {
    (0.10,1.05)
    (0.60,0.50)
    (1.05,1.50)
    (1.40,0.80)
    (1.85,1.25)
    (2.20,0.65)
    (2.55,1.40)
    (2.90,2.10)
    (3.35,0.75)
    (3.80,1.60)
    (4.15,2.00)
    (4.45,1.20)
};


\draw[
    purple!70!black,
    thick,
    smooth
]
plot coordinates {
    (0.15,0.35)
    (0.50,0.95)
    (1.00,0.80)
    (1.35,1.50)
    (1.80,0.45)
    (2.20,1.05)
    (2.50,0.70)
    (3.00,1.70)
    (3.35,0.45)
    (3.85,1.20)
    (4.20,0.85)
    (4.45,1.50)
};


\draw[orange!85!black, thick, smooth]
plot coordinates {
    (0.15,1.30)
    (0.60,1.85)
    (0.95,1.05)
    (1.30,2.00)
    (1.75,0.65)
    (2.15,1.15)
    (2.60,0.85)
    (3.05,1.55)
    (3.40,0.95)
    (3.85,1.95)
    (4.15,1.35)
    (4.45,2.15)
};


\fill[red] (1.80,2.10) circle (2pt);
\fill[red] (2.55,2.25) circle (2pt);
\fill[red] (2.90,2.10) circle (2pt);
\fill[red] (3.85,1.95) circle (2pt);
\fill[red] (4.45,2.15) circle (2pt);

\end{scope}

\node[
    font=\Large
] at ([xshift=0.05cm,yshift=-1.88cm]process.center)
{$t$};


\node[
    box,
    minimum width=2.5cm,
    minimum height=1.35cm,
    right=1.05cm of process,
    yshift=-1.5cm
] (tokmap)
{\large
\textbf{Tokenization}\\[1mm]
\large $A_t \in \{0,1,\ldots,d\}$
};

\draw[flow] (process.east) -| (tokmap.north);


\node[stage, minimum width=6.4cm, minimum height=2.2cm, below=1.05cm of tokmap] (tokens) {};

\node[font=\bfseries,anchor=north] at ([yshift=-0.15cm]tokens.north)
{\large Categorical token process};

\node[quiettoken]
    (t1)
    at ([xshift=-2.45cm,yshift=-0.3cm]tokens.center)
    {$0$};

\node[activetoken, right=0.16cm of t1]
    (t2) {$3$};

\node[quiettoken, right=0.16cm of t2]
    (t3) {$0$};

\node[activetoken, right=0.16cm of t3]
    (t4) {$5$};

\node[activetoken, right=0.16cm of t4]
    (t5) {$1$};

\node[activetoken, right=0.16cm of t5]
    (t6) {$4$};



\node[
    right=0.18cm of t6,
    font=\large
] {$\cdots$};


\draw[flow]
    (tokmap.south) -- (tokens.north);


\node[
    stage,
    minimum width=8.6cm,
    minimum height=3.8cm,
    right=0.9cm of tokmap,
    yshift=2.3cm
] (tr) {};

\node[
    font=\bfseries,
    anchor=north
] at ([yshift=-0.15cm]tr.north)
{\large Autoregressive transformer};


\coordinate (trin)  at ([xshift=-3.75cm,yshift=-0.30cm]tr.center);
\coordinate (trout) at ([xshift= 4.05cm,yshift=-0.30cm]tr.center);

\node[
    draw,
    rounded corners=4pt,
    thick,
    fill=purple!25,
    minimum width=0.90cm,
    minimum height=0.85cm,
    align=center
] (norm1) at ([xshift=-2.95cm,yshift=-0.30cm]tr.center) {\textrm{Norm}};

\node[
    draw,
    rounded corners=10pt,
    thick,
    fill=orange!30,
    minimum width=1.55cm,
    minimum height=1.45cm,
    align=center
] (attn) at ([xshift=-1.35cm,yshift=-0.30cm]tr.center)
{\large \textrm{MHSA}};

\node[
    draw,
    circle,
    thick,
    minimum size=0.55cm,
    inner sep=0pt,
    font=\Large
] (add1) at ([yshift=-0.30cm]tr.center) {$+$};

\node[
    draw,
    rounded corners=4pt,
    thick,
    fill=purple!25,
    minimum width=0.9cm,
    minimum height=0.85cm,
    align=center
] (norm2) at ([xshift=1.05cm,yshift=-0.30cm]tr.center) {\textrm{Norm}};

\node[
    draw,
    rounded corners=10pt,
    thick,
    fill=yellow!35,
    minimum width=1.55cm,
    minimum height=1.45cm,
    align=center
] (ffn) at ([xshift=2.60cm,yshift=-0.30cm]tr.center)
{\large \textrm{MLP}};

\node[
    draw,
    circle,
    thick,
    minimum size=0.55cm,
    inner sep=0pt,
    font=\Large
] (add2) at ([xshift=3.85cm,yshift=-0.30cm]tr.center) {$+$};

\draw[thick] (trin) -- (norm1.west);
\draw[thick] (norm1.east) -- (attn.west);
\draw[thick] (attn.east) -- (add1.west);
\draw[thick] (add1.east) -- (norm2.west);
\draw[thick] (norm2.east) -- (ffn.west);
\draw[thick] (ffn.east) -- (add2.west);
\draw[thick] (add2.east) -- (trout);

\draw[thick]
    (trin) |- ([xshift=-2.15cm,yshift=0.85cm]tr.center)
           -- ([yshift=0.85cm]tr.center)
           |- (add1.north);

\draw[thick]
    (add1.east) -- ++(0,0)
    coordinate (res2start);

\draw[thick]
    (add1.north) |- ([xshift=0.95cm,yshift=0.85cm]tr.center)
                 -- ([xshift=3.85cm,yshift=0.85cm]tr.center)
                 |- (add2.north);

\draw[flow]
    ([xshift=2.15cm]tokens.north)
    |- (trin);
    

\node[
    stage,
    minimum width=6.2cm,
    minimum height=4.3cm,
    right=-1.35cm of tr,
    yshift=-3.5cm
] (out) {};

\draw[flow]
    (trout.east) -| (out.north);

\node[
    font=\bfseries,
    align=center,
    anchor=north
] at ([yshift=-0.15cm]out.north)
{\large Estimated next-token\\
\large probabilities, $\widehat{\pib}_{t+1}$};

\coordinate (b0)
at ([xshift=-1.65cm,yshift=0.25cm]out.center);

\node[lab]
    at ($(b0)+(-1,0.52)$)
    {\large Quiet};

\draw[fill=blue!18]
    ($(b0)+(1.05,0.36)$)
    rectangle ++(2.95,0.18);

\node[lab]
    at ($(b0)+(-1.0,0.08)$)
    {\large Token 1};

\draw[fill=blue!18]
    ($(b0)+(1.05,-0.08)$)
    rectangle ++(0.60,0.18);

\node[lab]
    at ($(b0)+(-1.0,-0.36)$)
    {\large Token 2};

\draw[fill=blue!18]
    ($(b0)+(1.05,-0.52)$)
    rectangle ++(1.15,0.18);

\node[lab]
    at ($(b0)+(-1.0,-0.80)$)
    {\large Token 3};

\draw[fill=blue!18]
    ($(b0)+(1.05,-0.96)$)
    rectangle ++(0.82,0.18);

\node[lab]
    at ($(b0)+(-1.0,-1.24)$)
    {\large Token 4};

\draw[fill=blue!18]
    ($(b0)+(1.05,-1.40)$)
    rectangle ++(0.42,0.18);

\node[lab]
    at ($(b0)+(1.80,-1.68)$)
    {\huge $\mathbf{\vdots}$};


\node[
    box,
    font=\bfseries,
    minimum width=4.2cm,
    minimum height=0.95cm,
    below=0.55cm of out,
    align=center
] (sample)
{
\large Generate next token\\[1mm]
\large $A_{t+1} \sim \mathrm{Categorical} \! \left(\widehat{\pib}_{t+1}\right)$
};

\draw[flow]
    (out.south) -- (sample.north);

\end{tikzpicture}} \vspace{-0.5cm}
\caption{\footnotesize
Autoregressive transformer schematic based on the probabilistic setup in Section~\ref{subsec:tech}.}
\label{fig:pipeline}
\end{figure}
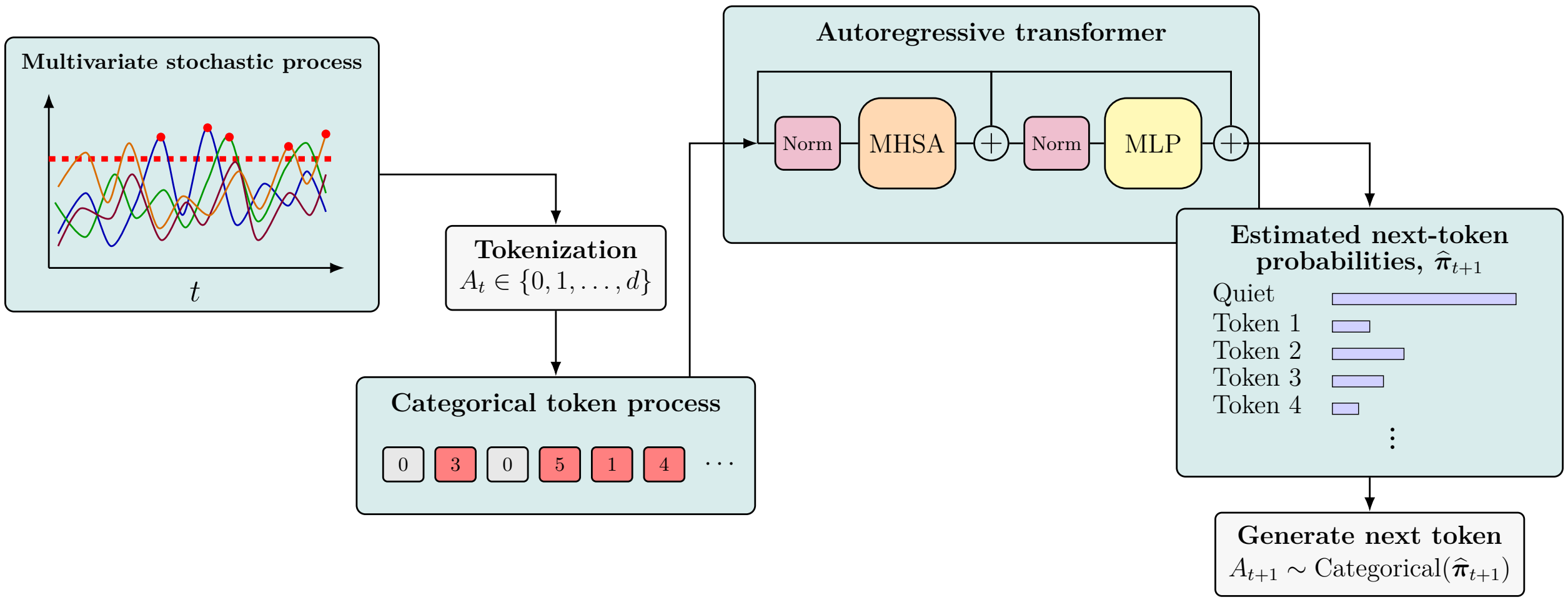

Let the window context at time $t$ be defined as
\begin{equation*}
    \mathcal{C}_t=(A_{s_t},\ldots,A_t),\qquad s_t=\max\{1,t-C+1\},
\end{equation*}
where $C\geq1$ is the maximum context length. Thus, $\mathcal{C}_t$ contains all available tokens when $t<C$ and the most recent $C$ tokens otherwise. The next token has conditional distribution
\begin{equation}\label{eq:next_token_distribution}
    \begin{aligned}
        A_{t+1}\mid\mathcal{C}_t
        &\sim\mathrm{Categorical}(\pib_{t+1}^{(q)}),\\
        \pib_{t+1}^{(q)}
        &=\bigl(\Pr\{A_{t+1}=0\mid\mathcal{C}_t\},\ldots,\Pr\{A_{t+1}=d\mid\mathcal{C}_t\}\bigr).
    \end{aligned}
\end{equation}
Below, dependence on the threshold level $q$ is suppressed whenever it is unambiguous and made explicit when studying the high-threshold limit. One of our key objectives is to estimate this distribution and use the estimate for the recursive generation of sequences of extremal tokens.
The quiet token in our setup is treated as an ordinary token, not an End-of-Sequence delimiter; it remains in $\mathcal{C}_t$ and neither terminates the sequence nor resets the conditioning history. 
Fig.~\ref{fig:pipeline} illustrates the tokenization scheme.

\subsubsection*{High-Threshold Limit of the Next-Token Kernel}
We next characterize the next-token kernel as the common marginal threshold level $q$ tends to one. Following common practice in multivariate extreme value theory, we first transform all margins to a common scale;
assuming continuous marginal distribution functions $F_j$, we transform the process $\{\bmY_t\}$ to standard Pareto margins via the probability integral transform,
\begin{equation*}
    Z_{t,j}=\frac{1}{1-F_j(Y_{t,j})},
    \qquad
    \bmZ_t=(Z_{t,1},\ldots,Z_{t,d})^{\T},
\end{equation*}
so that $\Pr(Z_{t,j}>z)=1/z$, for $z\geq1$. For each $q\in(0,1)$, the extremal token in \eqref{eq:token} then admits the equivalent representation
\begin{equation}\label{eq:token_pareto}
    A_t=\tau\left(\frac{\bmZ_t}{b(q)}\right),
\end{equation}
where $b(q) = 1/(1 - q)$, and 
\begin{equation*}
    \tau(\bmx)=
    \begin{cases}
        0, & \|\bmx\|_\infty\leq1,\\[1mm]
        \displaystyle\argmax_{1\leq j\leq d}x_j,
        & \|\bmx\|_\infty>1,
    \end{cases}
\end{equation*}
for $\|\cdot\|_\infty,$ the infinity norm. The $\argmax$ uses the same deterministic tie-breaking rule as \eqref{eq:token}.
The representation in \eqref{eq:token_pareto} follows because $Y_{t,j}>F_j^{-1}(q)$ if and only if $Z_{t,j}>b(q)$ almost surely, and $(1-u)^{-1}$ is strictly increasing for $u\in[0,1)$. Under the stationarity assumption below, for $t\geq C$ and whenever the conditioning event has positive probability, define the population next-token kernel, with its dependence on $q$ and the context made explicit, by
\begin{equation}\label{hight}
    \pi^{(q)}(a\mid\bmc)=\Pr\{A_{t+1}=a\mid\mathcal{C}_t=\bmc\},
\end{equation}
where $a\in\mathcal{V}$ is a candidate next extremal token and $\bmc=(c_1,\ldots,c_C)\in\mathcal{V}^C$ is an extremal token history. Our aim is to characterize \eqref{hight} directly as $q\to1$, or equivalently as $b(q)\to\infty$.

The \textit{unconditional} token process converges over every fixed time window to the all-quiet process because, as $q\to1$,
\begin{align*}
    \Pr\{A_t\neq0\}= \Pr\left\{\bigcup_{j=1}^d\left(Z_{t,j}>b(q)\right)\right\}\leq \sum_{j=1}^d\Pr\{Z_{t,j}>b(q)\}=\frac{d}{b(q)}=d(1-q)\to0.
\end{align*}

To characterize the high-threshold limit in \eqref{hight}, suppose that $\{\bmZ_t\}$ is stationary. For $m\in\{C,C+1\}$, assume that the block vector $(\bmZ_{t-m+1},\ldots,\bmZ_t)$ is multivariate regularly varying in the standard sense. Thus, for some nonzero Radon measure $\nu_m$ on $\mathbb{E}_m=([0,\infty]^d)^m\setminus\{\bm0\}$,
\begin{equation*}
    r\Pr\left\{r^{-1}(\bmZ_{t-m+1},\ldots,\bmZ_t)\in\,\cdot\,\right\}\xrightarrow{v}\nu_m(\,\cdot\,),\qquad r\to\infty,
\end{equation*}
where $\xrightarrow{v}$ denotes vague convergence \citep[Theorem~6.1]{resnick2007heavy}.

To express the limiting next-token kernel (see Theorem~\ref{thm:limiting_token_kernel}), for $\bmc\in\mathcal{V}^C$ and $a\in\mathcal{V}$, define the following sets of scaled Pareto blocks:
\begin{align*}
    B_{\bmc}
    &={}
    \left\{
        (\bmx_1,\ldots,\bmx_C)\in\mathbb{E}_C:
        \tau(\bmx_h)=c_h,\ h=1,\ldots,C
    \right\},\\
    B_{\bmc,a}
    &={}
    \left\{
        (\bmx_1,\ldots,\bmx_{C+1})\in\mathbb{E}_{C+1}:
        \tau(\bmx_h)=c_h,\ h=1,\ldots,C,\ 
        \tau(\bmx_{C+1})=a
    \right\}.
\end{align*}
For example, if $C=2$ and $\bmc=(0,2)$, then $B_{\bmc}$ consists of pairs $(\bmx_1,\bmx_2)\in\mathbb{E}_2$ for which $\bmx_1$ is quiet and component $2$ is the dominant exceedance in $\bmx_2$; $B_{\bmc,1}\subset\mathbb{E}_3$ additionally requires component $1$ to be the dominant exceedance in $\bmx_3$.

\begin{theorem}\label{thm:limiting_token_kernel}
Suppose that $\{\bmZ_t\}$ is stationary, $\bmc$ contains at least one non-quiet token, $0<\nu_C(B_{\bmc})<\infty$, and that $B_{\bmc}$ and $B_{\bmc,a}$, for every $a\in\mathcal{V}$, are continuity sets of $\nu_C$ and $\nu_{C+1}$, respectively. 
Then, as $q\to1$,
\begin{equation*}
    \pi^{(q)}(a\mid\bmc)\to
    \pi^{(1)}(a\mid\bmc)
    =\frac{\nu_{C+1}(B_{\bmc,a})}{\nu_C(B_{\bmc})},
    \qquad a\in\mathcal{V}.
\end{equation*}
Here, $\pi^{(1)}$ denotes the limiting kernel as $q\to1$. The limiting next-token kernel is nondegenerate, if and only if at least two next-token probabilities are positive, or equivalently,
    $\#\{a\in\mathcal{V}:\nu_{C+1}(B_{\bmc,a})>0\}\geq2$,
and it assigns positive probability to a non-quiet continuation if and only if $\sum_{a=1}^d\nu_{C+1}(B_{\bmc,a})>0$.
\end{theorem}

Theorem~\ref{thm:limiting_token_kernel} warrants some comments. First, when $C=1$, we identify the context $\bmc=(j)$ with its single token $j$. For $j\geq1$, the limiting lag-one conditional probability, from dominant component $j$ to token $a,$ is 
\begin{equation*}
    \pi^{(1)}(a\mid j)=\frac{\nu_2(B_{j,a})}{\nu_1(B_j)}.
\end{equation*}
In particular, $1-\pi^{(1)}(0\mid j)$ is the limiting probability that extreme activity remains present one step later. For $C>1$, the limit conditions on the complete recent token context. 
Second, to define the limiting law after recent extreme activity has disappeared, let $\bm0_C=(0,\ldots,0)\in\mathcal{V}^C$, and let $\mathcal{D}_C^+\subseteq\mathcal{V}^C\setminus\{\bm0_C\}$ be the set of length-$C$ contexts satisfying the context-specific hypotheses of Theorem~\ref{thm:limiting_token_kernel}. Extend the kernel by
\begin{equation*}
    \pi^{(1)}(\,\cdot\mid\bm0_C)=\delta_0,
\end{equation*}
where $\delta_0$ is the point mass at the quiet token, and set $\mathcal{D}_C=\mathcal{D}_C^+\cup\{\bm0_C\}\subseteq\mathcal{V}^C$. \appendixname~\ref{appdx:theory} verifies that this extension agrees with the above high-threshold limit.

\subsubsection*{Nonstationarity}
For continuous time-varying marginals, the construction extends by replacing $F_j$ and $u_j$ throughout with $F_{t,j}$ and $u_{t,j}=F_{t,j}^{-1}(q)$. The extremal vocabulary, common probability-scale comparison, and monotone-invariance property are preserved; dominant exceedances are then defined relative to the marginal distribution at time $t$. Theorem~\ref{thm:limiting_token_kernel} continues to apply if the standardized process $Z_{t,j}=\{1-F_{t,j}(Y_{t,j})\}^{-1}$ is stationary and satisfies the stated regular-variation and continuity conditions. If its extremal dependence also varies over time, however, the limiting tail measures become time-dependent, requiring a corresponding reformulation of the theorem; this more complex setting of nonstationary extremal dependence typically warrants separate treatment \citep[e.g.,][]{castro2018, decarvalho2024, murphy2025}.

\subsection{Autoregressive Transformer Architecture}\label{subsec:architecture}
An autoregressive transformer is a decoder-only transformer \citep{radford2018improving, bishop2024deep} that models the next-token probability vector $\pib_{t+1}$ in \eqref{eq:next_token_distribution} from the supplied context.

\paragraph{\textbf{Token embedding and positional encoding.}}
To process the discrete token sequence $\{A_t\}_{t=1}^n$ through an autoregressive transformer, each token $A_t\in\mathcal{V}$ is mapped to a continuous vector representing its identity and augmented by a positional encoding. Without positional information, unmasked self-attention is permutation-equivariant, so permuting the input positions permutes the corresponding outputs. Positional encodings identify locations within the supplied context, whereas the causal mask introduced below prevents a representation from using future tokens.

For a generic input segment $\{A_t\}_{t=1}^m$, each token $A_t$ is mapped to an embedding $\bme_t\in\mathbb{R}^{d_{\text{emb}}}$ and receives a learned absolute positional vector $\bme^{\text{pos}}_t\in\mathbb{R}^{d_{\text{emb}}}$, where $t=1,\ldots,m$. The resulting representation is $\bmx_t=\bme_t+\bme^{\text{pos}}_t$, and $\bmX=(\bmx_1,\ldots,\bmx_m)^{\T}\in\mathbb{R}^{m\times d_{\text{emb}}}$. Here $m\leq C$, with $C=256$ in all experiments, and the indices $1,\ldots,m$ refer to positions within the supplied segment rather than global time stamps or inferred stages of a cascade.

The cornerstone of the transformer architecture is the notion of self-attention \citep{bahdanau2014neural, vaswani2017attention}. The transformer is constructed by stacking $L$ transformer blocks with identical architectures, with each block containing its own learnable parameters and comprising two sublayers:
\begin{enumerate}[nolistsep,midpenalty=10000]
    \item masked multi-head self-attention (MHSA) layer;
    \item feed forward neural network layer.
\end{enumerate}
We describe a single transformer block, following the notation of \citet{vaswani2017attention} and \citet{bishop2024deep}. We use pre-normalization, in which layer normalization \citep{ba2016layer} is applied to the residual representation before each sublayer. Let $\bmX'=\mathrm{LayerNorm}(\bmX)\in\mathbb{R}^{m\times d_{\text{emb}}}$ be the pre-normalized representation of the input tokens.

\paragraph{\textbf{Sublayer 1: masked multi-head self-attention layer.}}
Assume that $H$ divides $d_{\text{emb}}$ and let $d_h=d_{\text{emb}}/H$. For each head $h\in\{1,\ldots,H\}$, let $\bmW_h^{(\text{q})},\bmW_h^{(\text{k})},\bmW_h^{(\text{v})}\in\mathbb{R}^{d_{\text{emb}}\times d_h}$ and $\bmb_h^{(\text{q})},\bmb_h^{(\text{k})},\bmb_h^{(\text{v})}\in\mathbb{R}^{d_h}$ be learnable projection matrices and biases, and let $\bm1_m\in\mathbb{R}^m$ be the vector of ones. The query, key, and value matrices are
\begin{equation*}
    \bmQ_h=\bmX'\bmW_h^{(\text{q})}+\bm1_m(\bmb_h^{(\text{q})})^{\T}, \qquad
    \bmK_h=\bmX'\bmW_h^{(\text{k})}+\bm1_m(\bmb_h^{(\text{k})})^{\T}, \qquad
    \bmV_h=\bmX'\bmW_h^{(\text{v})}+\bm1_m(\bmb_h^{(\text{v})})^{\T},
\end{equation*}
respectively, so that $\bmQ_h,\bmK_h,\bmV_h\in\mathbb{R}^{m\times d_h}$. The masked scaled dot-product self-attention for head $h$ is
\begin{equation*}
    \bmH_h=\mathrm{Attention}(\bmQ_h, \bmK_h,\bmV_h)=\mathrm{Softmax}\left(\frac{\bmQ_h\bmK_h^{\T}}{\sqrt{d_h}}+\bmM\right)\bmV_h,
\end{equation*}
where the softmax is applied row-wise and $\bmH_h\in\mathbb{R}^{m\times d_h}$. The causal mask $\bmM\in(\mathbb{R}\cup\{-\infty\})^{m\times m}$ has entries $M_{i,j}=0$ when $j\leq i$ and $M_{i,j}=-\infty$ when $j>i$, so the representation at position $i$ can attend only to positions $1,\ldots,i$. Division by $\sqrt{d_h}$ prevents the scale of the dot-product logits from growing with the head dimension.

Concatenating the head outputs gives $\bmH=\mathrm{Concat}[\bmH_1,\ldots,\bmH_H]\in\mathbb{R}^{m\times Hd_h}=\mathbb{R}^{m\times d_{\text{emb}}}$. With a learnable output projection $\bmW^{(o)}\in\mathbb{R}^{d_{\text{emb}}\times d_{\text{emb}}}$ and bias $\bmb^{(o)}\in\mathbb{R}^{d_{\text{emb}}}$, the multi-head self-attention output is
\begin{equation}\label{eq:multihead}
    \mathrm{MHSA}(\bmX';\bmM)=\mathrm{Concat}[\bmH_1,\ldots,\bmH_H]\bmW^{(o)}+\bm1_m(\bmb^{(o)})^{\T}\in\mathbb{R}^{m\times d_{\text{emb}}}.
\end{equation}
Concatenating the parallel heads preserves their distinct representations.

\paragraph{\textbf{Sublayer 2: feed forward neural layer.}}
Following the projection of the multi-head self-attention layer in \eqref{eq:multihead}, we again apply layer normalization \citep[\textrm{LayerNorm},][]{ba2016layer} and then pass the output into a multilayer perceptron (MLP) encased within a residual connection, which adds the input of the MLP directly to its output \citep{he2016deep}. This facilitates gradient propagation and generally improves optimization stability when stacking multiple transformer blocks later on.
Residual connections can help mitigate vanishing gradients, but do not guarantee that gradients remain nonzero. Thus, the neural layer follows as
\begin{equation*}
    \begin{split}
        \bmZ=\mathrm{MHSA}(\bmX';\bmM)+ \bmX, \qquad 
        \bmZ'=\mathrm{LayerNorm}(\bmZ), \qquad 
        \bmX^{\mathrm{out}}=\mathrm{MLP}(\bmZ')+\bmZ.
    \end{split}
\end{equation*}
The two layer-normalization operations within each block have distinct learned parameters. We iterate this forward pass across $L$ stacked transformer blocks with identical architecture. Further technical details on the transformer specification can be found in \appendixname~\ref{apdx:specs}.

\paragraph{\textbf{Final projection and probability mapping.}}
After the final transformer block, let $\widetilde{\bmX}=\mathrm{LayerNorm}_{f}(\bmX^{(L)})\in\mathbb{R}^{m\times d_{\text{emb}}}$. For the supplied context $\mathcal{C}_t$, whose length is $m$, the final row $\widetilde{\bmX}_{m,\cdot}\in\mathbb{R}^{d_{\text{emb}}}$ is projected onto the probability simplex to give the next-token distribution
\begin{equation} \label{eq:prob_mapping}
    \pib_{\thetab,t+1}=\pib_{\thetab}(\mathcal{C}_t) = \mathrm{Softmax}\left(\widetilde{\bmX}_{m,\cdot}\bmW^{(\mathrm{out})}+(\bmb^{(\mathrm{out})})^{\T}\right),
\end{equation}
where $\bmW^{(\mathrm{out})} \in \mathbb{R}^{d_{\text{emb}} \times |\mathcal{V}|}$ and $\bmb^{(\mathrm{out})}\in\mathbb{R}^{|\mathcal{V}|}$ are the learnable projection matrix and biases, respectively, and $\thetab$ collects all the estimable parameters of the decoder-only autoregressive transformer. The output in \eqref{eq:prob_mapping} is a parametrized model of the probability vector $\pib_{t+1}$ in \eqref{eq:next_token_distribution}.  The $j$th component is $\pi_{\thetab}(j\mid\mathcal{C}_t)=\Pr_{\thetab}(A_{t+1}=j\mid\mathcal{C}_t)$; for $j\in\{1,\ldots,d\}$, this is the conditional probability that component $j$ is selected as the dominant exceedance at time $t+1$, whereas $j=0$ gives the conditional probability of the quiet token. Accordingly, conditional on the first token, the finite-context model assigns
\begin{equation*}
    \Pr_{\thetab}(A_{2:n}\mid A_1)=\prod^{n-1}_{t=1}\Pr_{\thetab}(A_{t+1}\mid\mathcal{C}_t).
\end{equation*}
At position $t=1,\ldots,m$, the model predicts $A_{t+1}$ from the within-segment prefix $(A_1,\ldots,A_t)$. During training, $m+1$ consecutive tokens are extracted. The first $m$ are inputs and all $m$ outputs are paired with the targets $A_2,\ldots,A_{m+1}$. Thus, when a segment begins inside a longer trajectory, the context for its early positions is truncated at the segment boundary. At inference, the final position forecasts the token following the complete input segment. \appendixname~\ref{apdx:specs}, Algorithm~\ref{alg:trans}, gives the complete forward pass.

\subsection{Estimation and Amortization}\label{sec:amortization}
Estimation has two stages. First, each $F_j$ in \eqref{eq:token} is replaced by an estimator $\widehat F_j$ fitted using training observations only, yielding the token sequence $\{A_t\}_{t=1}^n$. The most natural option is the empirical distribution function, 
\begin{equation}\label{eq:empirical_cdf}
    \widehat{F}_j(y) = \frac{1}{n} \sum_{i=1}^n I(Y_{i,j} \leq y),
\end{equation}
which is the option taken below.\footnote{\appendixname~\ref{appdx:numexp} (Table~\ref{tab:mc_fixed_empirical_robustness}) compares alternatives to \eqref{eq:empirical_cdf}: a generalized Pareto model \citep[Chapter~4]{coles2001ismev} and time-varying kernel estimates \citep{harvey2012kernel}, under the same tokenization in \eqref{eq:token}.} Second, the transformer is fitted to estimate the probability vector in \eqref{eq:prob_mapping}; the transformer parameters $\thetab\in\Thetab$ are trained using categorical cross-entropy. Let $\mathcal{A}=(A_1,\ldots,A_{m+1})$ denote a sampled training segment, with fixed $1\leq m\leq C$ and positions reindexed locally. Its context at position $t$ is the prefix $\mathcal{C}_t=(A_1,\ldots,A_t)$. The average next-token negative log-likelihood scores all $m$ targets:
\begin{equation}\label{eq:seqlogloss}
    \mathcal{L}(\thetab;\mathcal{A})
    =-\frac1m\sum_{t=1}^{m}
      \log\left\{
        \pi_{\thetab}(A_{t+1}\mid\mathcal{C}_t)
      \right\}.
\end{equation}
The risk analysis uses the prediction kernel in \eqref{eq:prob_mapping} with dropout disabled; dropout is used during optimization. In the experiments, $m=C$.

Training on a single cascading trajectory may capture its specific features while providing limited coverage of other cascading regimes. Hence, we use amortized training \citep{ruoss2024amortized} across a parameterized family of data generating processes (DGPs) $\{F_{\psib}:\psib\in\Psi\}$. Here $F_{\psib}$ is the law of a sampled training segment under generative parameter $\psib$ and the training sampling protocol, including randomness in its trajectory location and tokenizer calibration.

Let $\Pi_{\psib}$ denote the training prior over DGP parameters. Following \citet{young2005essentials}, define the frequentist and Bayes risks of $\thetab$ for the loss in \eqref{eq:seqlogloss} as
\begin{equation*}
    \mathcal{R}_{\psib}(\thetab)
    =E_{\mathcal{A}\sim F_{\psib}}\{\mathcal{L}(\thetab;\mathcal{A})\},
    \qquad
    \mathcal{R}(\thetab)
    =E_{\psib\sim\Pi_{\psib}}\{\mathcal{R}_{\psib}(\thetab)\}.
\end{equation*}
The Bayes risk $\mathcal{R}(\thetab)$ characterizes the population prediction objective under the training hierarchy. 
Its empirical counterpart averages the loss over sampled training segments $\mathcal{D}=\{\mathcal{A}^{(1)},\ldots,\mathcal{A}^{(M)}\}$, which may share DGP parameters or overlap:
\begin{equation}\label{eq:empirical_amortized_risk}
    \widehat{\mathcal{R}}(\thetab) =\frac{1}{M}\sum_{i=1}^{M} \mathcal{L}(\thetab;\mathcal{A}^{(i)}).
\end{equation}
Training uses categorical cross-entropy with dropout to estimate a shared parameter $\widehat{\thetab} = \arg \min_{\thetab} \widehat{\mathcal{R}}(\thetab)$; \eqref{eq:empirical_amortized_risk} evaluates its prediction risk on the sampled segments.  The amortized parameters can be estimated by repeatedly refreshing the training dataset during optimization or at the end of every epoch. Details of amortized training are given with the simulation specification in \appendixname~\ref{appdx:numexp}. Importantly, the model parameters are not reinitialized; the same transformer weights are updated continuously.
We write $\widehat{\pib}(\bmc)=\pib_{\widehat{\thetab}}(\bmc)$ for the fitted next-token probability vector.

\subsection{Autoregressive Generation}\label{sec:autoregressive_generation}
For a parameter vector $\thetab$, let $\pi_{\thetab}(a\mid\bmc)$ denote the probability assigned to token $a\in\mathcal{V}$ by the transformer given context $\bmc$, as defined in \eqref{eq:prob_mapping}. The fitted kernel is $\pi_{\widehat{\thetab}}$. Recursively sampling from this kernel defines an ancestral sampler for the corresponding joint predictive law \citep[Section~12.3.2]{bishop2024deep}. Let $\widehat{\pib}_{t+1}=\widehat{\pib}(\mathcal{C}_t)$ be the fitted transformer's output based on the observed context $\mathcal{C}_t$. This fitted vector estimates $\pi^{(q)}(\,\cdot\mid\mathcal{C}_t)$ for fixed exceedance level $q\in(0,1)$ close to one. Its $j$th component is
\begin{equation*}
    \widehat{\pi}_{t+1,j}
    =\pi_{\widehat{\thetab}}(j\mid\mathcal{C}_t)
    =\Pr_{\widehat{\thetab}}\!\left(A_{t+1}=j\mid\mathcal{C}_t\right),
    \qquad j\in\mathcal{V}.
\end{equation*}
The first generated token is drawn according to
\begin{equation*}
    A^{\mathrm{gen}}_{t+1}\mid\mathcal{C}_t
    \sim \mathrm{Categorical}\!\left(\widehat{\pib}_{t+1}\right).
\end{equation*}
After each draw, the generated token is appended to the context and, if its length exceeds $C$, the oldest token is removed. Let $\mathcal{C}^{\mathrm{gen}}_{t+h-1}$ be this updated context before the $h$th draw, with $\mathcal{C}^{\mathrm{gen}}_t=\mathcal{C}_t$. Recursively,
\begin{equation*}
    A^{\mathrm{gen}}_{t+h}\mid\mathcal{C}^{\mathrm{gen}}_{t+h-1}
    \sim \mathrm{Categorical}\!\left(\widehat{\pib}_{t+h}\right),
    \qquad h=1,\ldots,H,
\end{equation*}
where $\widehat{\pib}_{t+h}=\widehat{\pib}(\mathcal{C}^{\mathrm{gen}}_{t+h-1})$. These recursive draws define the $H$-step rollout distribution, that is, the joint distribution of the complete generated continuation $\mathbf{A}^{\mathrm{gen}}_{t+1:t+H}=(A^{\mathrm{gen}}_{t+1},\ldots,A^{\mathrm{gen}}_{t+H})$, rather than only that of its final token. To write this distribution explicitly, fix a candidate continuation $\mathbf{a}_{t+1:t+H}=(a_{t+1},\ldots,a_{t+H})$, and let $\bmc_{t+h-1}$ be the context obtained by appending $a_{t+1},\ldots,a_{t+h-1}$ to $\mathcal{C}_t$ and retaining the most recent $C$ tokens. Its probability under the fitted kernel is
\begin{equation*}
    \Pr_{\widehat{\thetab}} \left(\mathbf{A}^{\mathrm{gen}}_{t+1:t+H}=\mathbf{a}_{t+1:t+H}\mid\mathcal{C}_t\right)=\prod_{h=1}^{H}\Pr_{\widehat{\thetab}} \left(A^{\mathrm{gen}}_{t+h}=a_{t+h}\mid\bmc_{t+h-1}\right).
\end{equation*}
Thus, at $T=1$, each generated trajectory is one draw from the fitted rollout distribution. A generated quiet token, $A^{\mathrm{gen}}_{t+h}=0$, is appended to the context like any other token and does not stop the recursion.

Below, let $P_{\thetab,\bmc}^{(H)}$ be the distribution on $\mathcal{V}^H$ obtained by recursively applying $\pi_{\thetab}$, and $P_{1,\bmc}^{(H)}$ be the distribution obtained from the limiting kernel $\pi^{(1)}$ in Theorem~\ref{thm:limiting_token_kernel}, for an initial context $\bmc\in\mathcal{D}_C$ and horizon $H\geq1$.  

The following result compares these rollout distributions using the total variation distance over the contexts reached by the limiting rollout. Recall that for probability measures $P$ and $Q$ on the same finite set $\mathcal{X}$, their total variation distance is
\begin{equation*}
    \text{d}_{\mathrm{TV}}(P,Q)
    =\sup_{B\subseteq\mathcal{X}}|P(B)-Q(B)|
    =\frac12\sum_{x\in\mathcal{X}}|P(\{x\})-Q(\{x\})|.
\end{equation*}
Here, $\mathcal{X}=\mathcal{V}$ for next-token distributions and $\mathcal{X}=\mathcal{V}^H$ for rollout distributions. 

\begin{theorem}\label{thm:limiting_generated_law}
Fix an initial context $\bmc\in\mathcal{D}_C$ and an integer $H\geq1$. Let $\mathcal{S}_H := \mathcal{S}_H(\bmc)$ be the set of contexts that can occur with positive probability immediately before one of the first $H$ draws in a rollout obtained by recursively applying $\pi^{(1)}$ from $\bmc$, including $\bmc$ itself before the first draw, and suppose that $\mathcal{S}_H \subseteq\mathcal{D}_C$. For any transformer kernel $\pi_{\thetab}$ on $\mathcal{V}^C$, define
\begin{equation*}
    \delta_{\thetab}=
    \sup_{\bmr\in\mathcal{S}_H(\bmc)}
    \text{\upshape d}_{\mathrm{TV}}\{\pi_{\thetab}(\,\cdot\mid\bmr),
    \pi^{(1)}(\,\cdot\mid\bmr)\}.
\end{equation*}
Then
\begin{equation*}
    \text{\upshape d}_{\mathrm{TV}}\{P_{\thetab,\bmc}^{(H)},P_{1,\bmc}^{(H)}\}
    \leq1-(1-\delta_{\thetab})^H
    \leq H\delta_{\thetab}.
\end{equation*}
Thus, for fixed $H$, any sequence $\{\thetab_\ell\}$ satisfying $\delta_{\thetab_\ell}\to0$ also satisfies $P_{\thetab_\ell,\bmc}^{(H)}\to P_{1,\bmc}^{(H)}$ in total variation.
\end{theorem}

Theorem~\ref{thm:limiting_generated_law} shows that uniform one-step accuracy over the contexts encountered by the limiting rollout guarantees accuracy of the joint distribution of the entire $H$-token continuation. The rollout error is bounded by at most $H$ times the largest one-step error. In words, a key takeaway of Theorem~\ref{thm:limiting_generated_law} is that accurate next-token probabilities yield an accurate generated path distribution.

To separate learning error from the high-threshold approximation, write $\mathcal{S}_H=\mathcal{S}_H(\bmc)$ and, for $q$ sufficiently close to one that the relevant finite-threshold kernels are defined, let
\begin{equation*}
    \delta_{\thetab,q}^{\mathrm{fit}}=
    \sup_{\bmr\in\mathcal{S}_H}
    \text{d}_{\mathrm{TV}}\{\pi_{\thetab}(\,\cdot\mid\bmr),\pi^{(q)}(\,\cdot\mid\bmr)\},\qquad
    \delta_q^{\mathrm{thr}}=
    \sup_{\bmr\in\mathcal{S}_H}
    \text{d}_{\mathrm{TV}}\{\pi^{(q)}(\,\cdot\mid\bmr),\pi^{(1)}(\,\cdot\mid\bmr)\}.
\end{equation*}
Here, $\delta_{\thetab,q}^{\mathrm{fit}}$ is the one-step fitting error at threshold $q$ and $\delta_q^{\mathrm{thr}}$ is the threshold approximation error. The triangle inequality gives
\begin{equation*}
    \delta_{\thetab}
    \leq
    {\delta_{\thetab,q}^{\mathrm{fit}}}
    +
    {\delta_q^{\mathrm{thr}}}.
\end{equation*}
Because $\mathcal{S}_H$ is finite, Theorem~\ref{thm:limiting_token_kernel} and its all-quiet extension imply $\delta_q^{\mathrm{thr}}\to0$ as $q\to1$. Hence, along any sequence $q_\ell\to1$ and fitted kernels $\pi_{\thetab_\ell}$ for which $\delta_{\thetab_\ell,q_\ell}^{\mathrm{fit}}\to0$, the corresponding fixed-horizon rollout laws converge to $P_{1,\bmc}^{(H)}$ in total variation.

\subsection{Temperature-Adjusted Generation}\label{sec:temperature_generation}
As is standard \citep[e.g.,][Chapter~12]{bishop2024deep}, temperature-adjusted generation can be performed by dividing the softmax logits by a temperature $T>0$; the unadjusted case $T=1$ corresponds to generation based on \eqref{eq:prob_mapping}, for which the rollout-law interpretation and the stability result in Theorem~\ref{thm:limiting_generated_law} apply.

For $T\neq1$, temperature scaling defines a different next-token kernel, so the theorem does not by itself guarantee fidelity to $\pi^{(1)}$. Despite this, in applications such as finance, exploring alternative cascade patterns, including unusual token transitions or unexpected propagation of extreme losses, may motivate sampling with $T\neq1$.

Extending \eqref{eq:prob_mapping}, temperature-adjusted conditional probabilities can be obtained as
\begin{equation*} 
    \widehat{\pib}^{(T)}_{t+1} = \mathrm{Softmax}\left(\frac{\widetilde{\bmX}_{m,\cdot}\bmW^{(\mathrm{out})}+(\bmb^{(\mathrm{out})})^{\T}}{T}\right).
\end{equation*}
If the maximizing probability is unique, the distribution concentrates on that token as $T\to0$; if several probabilities share the maximum, the limiting distribution is uniform over those tokens. When $T>1$, the probability mass is spread out more evenly across possible tokens, increasing the likelihood of sampling less probable outcomes. 
We further investigate temperature-adjusted generation of extremal tokens in Section~\ref{subsec:financial_temperature}.

\section{Cascading Extreme Returns in Semiconductor Stocks}\label{sec:application}
\subsection{Motivation, Data and Transfer Learning Design}
\paragraph{\textbf{Semiconductor stocks data.}} Fueled by the AI infrastructure boom, the importance of the semiconductor industry within the tech sector has substantially increased \citep{haramboure2023vulnerabilities}. These firms are also strongly interconnected through common market cycles, supply chains, and exposure to macroeconomic and major technology sector shocks. This provides a natural real-world setting for studying cascading extremes and, in particular, for assessing whether the proposed transformer can learn the conditional propagation of extreme losses in high-frequency stock-price movements. 

Data are observed at ten-second intervals and consist of the five largest constituents of the Philadelphia Semiconductor Index (SOX) by market capitalization: NVIDIA Corp. (\textrm{NVDA}, \$5.01~Tr), Taiwan Semiconductor Manufacturing Company Ltd. (\textrm{TSM}, \$2.09~Tr), Broadcom Inc. (\textrm{AVGO}, \$1.82~Tr), Micron Technology Inc. (\textrm{MU}, \$1.04~Tr), and Advanced Micro Devices Inc. (\textrm{AMD}, \$0.85~Tr). The raw data span 2nd January 2020 to 24th July 2026, contain $1{\,}648$ trading days, and were obtained from the Alpaca Market Data API \citep{alpaca2026api} using the Securities Information Processor feed\footnote{A centralized system that aggregates all-inclusive trade and quote data from every U.S. exchange forming the official National Best Bid and Offer (NBBO).}. 
The observation period includes several significant market stresses, including the COVID-19 market crash in March 2020, the 2025 tariff crash, and an oil-price spike in spring 2026.\footnote{\appendixname~\ref{appdx:eda} (Fig.~\ref{fig:intraday}) illustrates persistent and recurrent dominant-exceedance activity on 23rd March 2020. Repeated transitions among stocks during market-wide stress motivate a propagation model rather than isolated exceedance prediction.}

We retain stock prices from 09:00 to 16:00 Eastern Time, an interval that includes the final 30 minutes of premarket trading. These observations allow the token sequence to capture extreme movements immediately preceding the regular trading session. A missing premarket price often indicates that no trade occurred; we treat such periods as \textrm{Quiet} tokens. We therefore forward fill the most recently recorded price, which yields a zero negative log return.

\paragraph{\textbf{Data preparation and evaluation setup.}} Since we focus on extreme losses, our units of analysis below are negative returns. 
For stock $j$ with price $P_{t,j}$ at time $t$, we define the negative log return as
\begin{equation*}
    Y_{t,j}=-\log\left(\frac{P_{t,j}}{P_{t-1,j}}\right),
\end{equation*}
so that larger positive values correspond to larger proportional price declines. At each time point, the five loss returns are mapped to the extremal vocabulary
\begin{equation}
  \label{eq:vv}
  \mathcal{V}\!=\!\{\mathrm{Quiet}, \mathrm{NVDA},\mathrm{TSM}, \mathrm{AVGO}, \mathrm{MU}, \mathrm{AMD}\},  
\end{equation}
using \eqref{eq:token}. Thus, whenever the token is non-quiet, its ticker identifies the stock selected as the dominant exceedance at that time. 

Observations from 2nd January 2020 through 31st March 2026 constitute the training period, and those from 1st April through 31st May 2026 form the validation period, comprising $1\,569$ days ($3\,955\,449$ tokens) and $41$ days ($103\,361$ tokens), respectively. The remaining observations through 24th July 2026 are reserved for testing and comprise $38$ trading days and $86\,031$ data points. We set $q=0.98$ as it retains enough extremal tokens for evaluating cascading behavior and matches the simulation setup from \appendixname~\ref{appdx:numexp}.

We apply the pretrained model from \appendixname~\ref{sec:simul} to the extremal token sequence and compare four levels of adaptation \citep{radford2018improving,raffel2020exploring,yenduri2024gpt}.

\paragraph{\textbf{Transfer learning and evaluation.}} Zero-shot uses the pretrained model without adaptation; one-shot and few-shot use one and five epochs of adaptation, respectively, to the training data. We run adaptation for $200$ epochs, and checkpoint the optimal-shot when the lowest validation cross-entropy is attained. To illustrate the numerical benefits of pretraining, the four pretrained variants are compared against a \textit{non-pretrained} benchmark model. Model configurations follow Appendices~\ref{appdx:numexp} and~\ref{apdx:specs}.

Generation uses $1\,000$ Monte Carlo repetitions and context length of $256$ tokens. Test rollouts start from the final validation context. We compare the observed and generated continuations following the same test contexts. We assess predictions and generated trajectories against empirical test period token frequencies, transition probabilities, and persistence, since the true next-token kernel is unknown.

\subsection{Empirical Results}
\begin{figure}[!htbp]
\centering \vspace{-0.3cm}
\begin{minipage}[c]{0.47\textwidth}
\centering
\includegraphics[width=\linewidth, trim=0 0 0 0, clip]{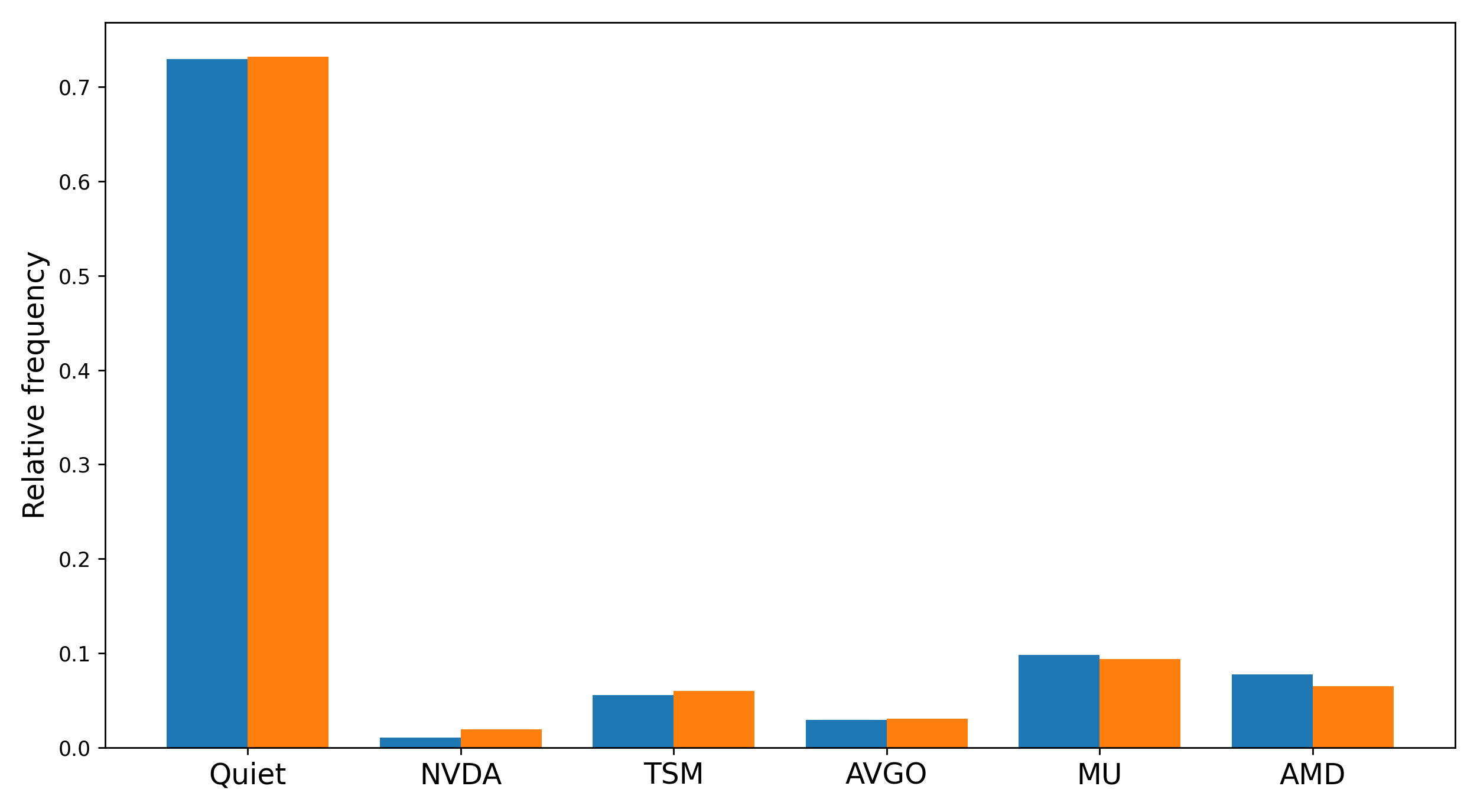}
\end{minipage}
\hfill
\begin{minipage}[c]{0.52\textwidth}
\centering
\scriptsize
\setlength\extrarowheight{4pt}
{\textrm{Table} 1}\label{tab:pl}\\
\vspace{0.1cm}
\resizebox{\textwidth}{!}{
\begin{tabular}{c|cccccc|c}
\toprule
& \textrm{Quiet} & \textrm{NVDA} & \textrm{TSM} & \textrm{AVGO} & \textrm{MU} & \textrm{AMD} & $\text{d}_{\mathrm{TV}}$ \\
\hline
Observed
& $0.729$ & $0.011$ & $0.055$ & $0.029$ & $0.098$ & $0.077$ & -\\
\hline
Zero-shot
& $0.818$ & $0.020$ & $0.036$ & $\mathbf{0.027}$ & $0.056$ & $0.043$ & $0.098$ \\
One-shot
& $0.795$ & $0.022$ & $0.033$ & $0.027$ & $0.057$ & $0.064$ & $0.078$ \\
Few-shot
& $0.705$ & $0.018$ & $0.068$ & $0.036$ & $0.106$ & $0.066$ & $0.035$ \\
Optimal-shot
& $\mathbf{0.739}$ & $\mathbf{0.018}$ & $\mathbf{0.056}$ & $0.032$ & $\mathbf{0.091}$ & $0.064$ & $\mathbf{0.021}$ \\
Non-pretrained
& $0.762$ & $0.025$ & $0.037$ & $0.038$ & $0.072$ & $\mathbf{0.067}$ & $0.055$ \\
\bottomrule
\end{tabular}}
\end{minipage}
\caption{\footnotesize Left: Relative frequency histograms comparing observed marginal token frequencies with those generated by the optimal-shot model at $T=1$. Right: \tablename~1 reports the marginal probabilities and the total variation distance at each stage of transfer learning and the non-pretrained model. Values in bold are the marginal probabilities closest in absolute distance to the corresponding observed empirical values, or the lowest estimates of the total variation distance.}\label{fig:marginals}
\end{figure}

\paragraph{\textbf{Predictive performance.}}
Fig.~\ref{fig:marginals} compares the observed marginal token distribution with Monte Carlo generation from the optimal-shot model obtained at epoch $17$ with temperature $T\!=\!1$. The main deviation is for the \textrm{Quiet} token, to which the model assigns probability $0.739$ compared with $0.729$ empirically, indicating a close agreement to predict termination of ongoing extreme activity. The model captures the allocation of dominant-exceedance probability across stocks well, with particularly accurate marginal estimates for \textrm{MU} and \textrm{TSM} and a maximum discrepancy of $0.013$ for \textrm{AMD}.

\begin{figure}[!htbp]
    \centering
    \includegraphics[width=0.99\linewidth]{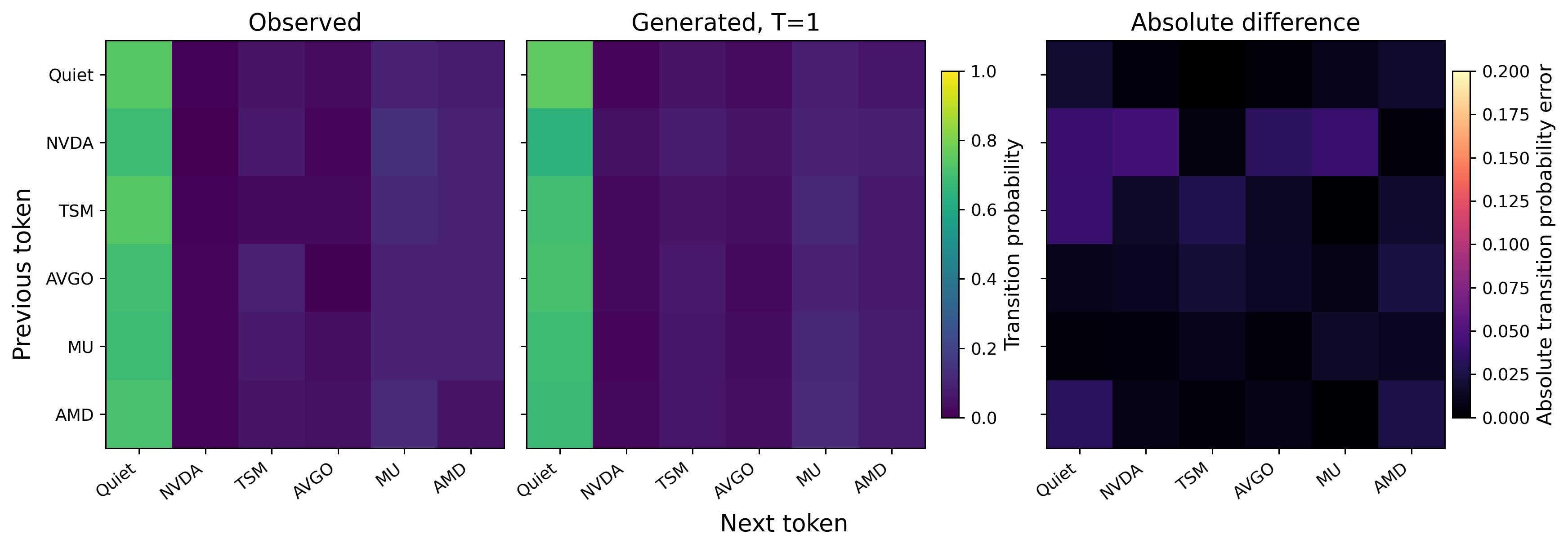}
    \vspace{-0.5cm}
    \caption{\footnotesize Empirical (left) and optimal-shot model (centre) transition probabilities, with tilewise absolute differences (right).}
    \label{fig:fewheatmap}
\end{figure}

To assess sequential structure, \figurename.~\ref{fig:fewheatmap} compares transition frequencies pooled across matched observed and generated continuations. Each generated continuation starts from the same observed test history as its matched observed continuation. For tokens $a,c\in\mathcal{V}$ in \eqref{eq:vv}, let $N^{\mathrm{obs}}_{c,a}$ and $N^{\mathrm{gen}}_{c,a}$ count transitions from $c$ to $a$ within these continuations, pooled separately over the observed and generated samples. The transition probability estimates are
\begin{equation*}
    \widehat{\pi}^{\mathrm{obs}}(a\mid c)
    =\frac{N^{\mathrm{obs}}_{c,a}}
    {\sum_{b\in\mathcal{V}}N^{\mathrm{obs}}_{c,b}},
    \qquad
    \widehat{\pi}^{\mathrm{gen}}(a\mid c)
    =\frac{N^{\mathrm{gen}}_{c,a}}
    {\sum_{b\in\mathcal{V}}N^{\mathrm{gen}}_{c,b}}.
\end{equation*}
Fig.~\ref{fig:fewheatmap} reports the entries and the absolute differences $\lvert\widehat{\pi}^{\mathrm{gen}}(a\mid c)-\widehat{\pi}^{\mathrm{obs}}(a\mid c)\rvert$, for the optimal-shot model at $T=1$. Agreement indicates that the generated continuations reproduce the empirical one-step transition structure, including stock-specific dominant exceedances. These are pooled lag-one frequencies, not the transformer's full-context predictive kernel.
\vspace{-0.3cm}
\vspace{-0.5cm}
The transfer-learning path is consistent with a bias--variance tradeoff. Zero-shot over-assigns probability to the \textrm{Quiet} token and under-represents several dominant-exceedance tokens, whereas optimal-shot reduces marginal total variation distance from $0.098$ to $0.021$ and gives the smallest transition discrepancies in Fig.~\ref{fig:transition} (second row). The optimal-shot also outperforms the non-pretrained model, suggesting that our judicious choice of initialisation helps to improve training and, thus, the predictive performance of our generator.

\subsection{Temperature-Adjusted Monte Carlo Generation}\label{subsec:financial_temperature}
Following Section~\ref{sec:temperature_generation}, we compare generation across temperature levels $T\in\{0.95,1,1.1\}$. The complete transition-error grid is reported in \appendixname~\ref{appdx:eda} (Fig.~\ref{fig:transition}). At $T=1$, the transition discrepancies are generally small, particularly for the optimal-shot model, indicating that recursively generated paths preserve the empirical balance between cascade termination and continuation without adjustment. Departures from the default temperature alter this balance substantially. Lowering the temperature to $T=0.95$ reduces transition errors for the few-shot specification and retains relatively small discrepancies for the optimal-shot model, while larger deviations remain for the less-adapted specifications, particularly in transitions involving the \textrm{Quiet} state. In contrast, increasing the temperature to $T=1.1$ produces pronounced errors in transitions into \textrm{Quiet} for the few-shot and optimal-shot models, indicating that greater sampling dispersion could distort the conditional distributions. Thus, temperature tuning can generate scenario-specific token trajectories and induce systematic changes.

Overall, $T=1$ is the natural unadjusted benchmark, under which the optimal-shot model most closely represents the observed transition dynamics. Comparison across adaptation stages further shows that the pretrained model already contains a useful dependence structure for cascading extremes; financial adaptation primarily refines the relative probabilities of token-specific propagation and termination rather than learning the transition structure from scratch.

\section{Discussion}\label{sec:conclusion}
This paper introduces generative models for extremal tokens, providing a natural interface between extreme value theory and the autoregressive architectures underlying large language models. The token representation retains the identity and temporal ordering of dominant exceedances without modeling their magnitudes. Our theoretical results characterize the high-threshold next-token law and bound fixed-horizon path-distribution error in terms of uniform one-step approximation error. The simulations support conditional-probability estimation across the studied process families, while the financial application illustrates how modest adaptation can reproduce observed cascade patterns. While cascades can, in principle, also be generated from a fitted parametric model, for either continuous or discrete series (e.g., the fitted VAR($p$) model in \appendixname~\ref{appdx:eda}), our transformer-based approach targets conditional extremal token probabilities directly, preserving invariance to strictly increasing marginal transformations and potentially reducing sensitivity to misspecification of the return dynamics.

Some comments on future research are in order. Firstly, it would seem natural to extend the proposed setting to generate transitions of dominant exceedances or full exceedance rankings jointly with latent variables, such as bull and bear market regimes. Secondly, while the focus here has been on dominant exceedances, in practice one may also be interested in constructing a richer vocabulary that records the ranking of all exceedances. Tokens could then be ordered tuples $(j_1,\ldots,j_k)$ of distinct exceeding components, ranked in decreasing order of their marginal probability-scale scores, with the empty tuple representing quiet observations. For five components, there are $\binom{5}{k}$ possible subsets of $k$ exceeding components and $k!$ rankings of each subset. Including the quiet token, the vocabulary therefore contains $$1+\sum_{k=1}^{5}\binom{5}{k}k!=326~\text{tokens},$$ rather than $6$ as in this paper. While natural, such extension would be computationally much more intensive as the ranked-exceedance vocabulary yields $O(d!d_{\text{emb}})$ complexity and would call for an extension of Theorem~\ref{thm:limiting_token_kernel}. Thirdly, in practice one may also be interested in the actual magnitudes of extremes, which would require an approach grounded in EVT. Combining transformer-based architectures with multivariate extreme value theory is outside the scope of this work but would be a natural extension. Theorem~\ref{thm:limiting_token_kernel} could perhaps provide some hints on how this could be achieved. We leave such open problems for future analysis.

\section*{Acknowledgments}
This work is supported by the Centre for Investing Innovation of the Univeristy of Edinburgh. We also thank the participants of GAME 2026 (Generative AI Modeling for Extreme Events), GL(EN)$^2$ (the Glasgow--Lancaster--Edinburgh--Newcastle Extremes Network), GAIL (UoE Generative AI Lab), ECFI (Edinburgh Centre for Financial Innovations), Nikolaos Makras and Gr\'egoire Clart\'e for insightful discussions. This work has made use of the computing resources provided by the Edinburgh Compute and Data Facility. 

\bibliographystyle{tmlr}
\bibliography{ref.bib}

\appendix
\raggedbottom
\setlength{\floatsep}{6pt plus 2pt minus 1pt}
\setlength{\textfloatsep}{8pt plus 2pt minus 1pt}
\setlength{\intextsep}{6pt plus 2pt minus 1pt}
\makeatletter
\setlength{\@fptop}{0pt}
\setlength{\@fpsep}{6pt}
\setlength{\@fpbot}{0pt plus 1fil}
\makeatother

\section{Proofs and Technical Details}\label{appdx:theory}

\subsection{Proofs of the Main Results}\label{appdx:proofs}
This subsection verifies monotone invariance, proves Theorems~\ref{thm:limiting_token_kernel} and~\ref{thm:limiting_generated_law}, and establishes the all-quiet extension noted in Section~\ref{subsec:tech}. The kernel limits are taken as $q\to1$ under the notation and assumptions of Section~\ref{subsec:tech}, with $C$ fixed. 

\textit{Monotone invariance.}
Let $\widetilde Y_{t,j}=g_j(Y_{t,j})$, where each $g_j$ is strictly increasing, and let $\widetilde F_j$ be its marginal distribution function and $\widetilde u_j=g_j(u_j)$. Then
\begin{equation*}
    \widetilde F_j\{g_j(y)\}
    =\Pr\{g_j(Y_{t,j})\leq g_j(y)\}=F_j(y),
    \qquad
    I(\widetilde Y_{t,j}>\widetilde u_j)=I(Y_{t,j}>u_j),
\end{equation*}
from where it immediately follows that for every $t$,
\begin{equation*}
    \widetilde A_t=A_t,
\end{equation*}
where $\widetilde A_t$ is obtained from \eqref{eq:token} using the transformed observations, marginals, and thresholds; that is, componentwise strictly increasing transformations leave the extremal token sequence unchanged.
\begin{proof}[Proof of Theorem~\ref{thm:limiting_token_kernel}]
Let
\begin{equation*}
    D_q=\Pr\{\mathcal{C}_t=\bmc\},
    \qquad
    N_{q,a}=\Pr\{\mathcal{C}_t=\bmc,
    A_{t+1}=a\}.
\end{equation*}
The representation $A_s=\tau\left(\bmZ_s/b(q)\right)$ gives
\begin{align*}
    D_q
    =\Pr\!\left\{
        b(q)^{-1}(\bmZ_{t-C+1},\ldots,\bmZ_t)
        \in B_{\bmc}
    \right\}, \qquad 
    N_{q,a}
    =\Pr\!\left\{
        b(q)^{-1}(\bmZ_{t-C+1},\ldots,\bmZ_{t+1})
        \in B_{\bmc,a}
    \right\}.
\end{align*}
Because $\bmc$ contains a non-quiet token, every block in either set has at least one vector with sup-norm greater than one. Thus, both sets are bounded away from the origin and relatively compact in the compactified cone. Vague convergence, the continuity-set assumptions, and stationarity for the $(C+1)$-block ending at $t+1$ therefore yield that, as $q\to1$, 
\begin{equation*}
    b(q)D_q\to\nu_C(B_{\bmc}),
    \qquad
    b(q)N_{q,a}\to\nu_{C+1}(B_{\bmc,a}),
\end{equation*}
Since $\nu_C(B_{\bmc})>0$, we have $D_q>0$ for $q$ sufficiently close to one, and hence
\begin{equation*}
    \pi^{(q)}(a\mid\bmc)
    =\frac{N_{q,a}}{D_q}
    \to
    \frac{\nu_{C+1}(B_{\bmc,a})}
         {\nu_C(B_{\bmc})}.
\end{equation*}
For every $q$, the events defining $N_{q,a}$, $a\in\mathcal{V}$, form a disjoint partition of the event defining $D_q$. Hence $D_q=\sum_{a\in\mathcal{V}}N_{q,a}$. Multiplying by $b(q)$ and taking limits, with the finite sum passing through the limit, gives
\begin{equation*}
    \nu_C(B_{\bmc})
    =\sum_{a\in\mathcal{V}}\nu_{C+1}(B_{\bmc,a}),
\end{equation*}
so the limiting ratios define the categorical next-token distribution $\pi^{(1)}(\,\cdot\mid\bmc)$. This distribution is nondegenerate exactly when at least two next-token probabilities are positive, and it assigns positive probability to a non-quiet continuation exactly when $\sum_{a=1}^d\nu_{C+1}(B_{\bmc,a})>0$.
\end{proof}

\noindent\textit{Proof of all-quiet extension of Theorem~\ref{thm:limiting_token_kernel}.}
For fixed $C$, stationarity and the same union-bound argument for the unconditional token process yield, as $q\to1$,
\begin{equation*}
    \Pr\{\mathcal{C}_t=\bm0_C\}
    \geq 1-\frac{Cd}{b(q)}
    \to1,
    \qquad
    \Pr\{A_{t+1}\neq0\}
    \leq\frac{d}{b(q)}
    \to0.
\end{equation*}
Hence the all-quiet extension stated in Section~\ref{subsec:tech} agrees with the one-step high-threshold limit because
\begin{equation*}
    1-\pi^{(q)}(0\mid\bm0_C)
    \leq
    \frac{\Pr\{A_{t+1}\neq0\}}
         {\Pr\{\mathcal{C}_t=\bm0_C\}}
    \to0. 
\end{equation*}
Consequently, for every fixed $H$, the probability that a rollout obtained by recursively applying $\pi^{(q)}$ from $\bm0_C$ contains a non-quiet token equals
\begin{equation*}
    1-\{\pi^{(q)}(0\mid\bm0_C)\}^H
    \leq H\{1-\pi^{(q)}(0\mid\bm0_C)\}
    \to0.
\end{equation*}
Thus the fixed-horizon rollout law converges to the point mass on the all-quiet path. \qed

\begin{proof}[Proof of Theorem~\ref{thm:limiting_generated_law}]
For a fixed $\thetab$, couple a rollout from $\pi_{\thetab}$ and a rollout from $\pi^{(1)}$, both initialized at $\bmc$, by using a maximal coupling at each step whenever their generated histories agree and any coupling after they differ. Let $\mathcal{G}_h$ be the sigma-field generated by the coupled rollouts through step $h$, and let $E_h$ be the event that their first $h$ generated tokens agree, with $E_0$ certain. On $E_{h-1}$, the two rollouts have the same context $\bmr_h$, which belongs to $\mathcal{S}_H(\bmc)$ because it is also the current context of the limiting rollout. Their next tokens can therefore be coupled to agree with conditional probability at least
\begin{equation*}
    1-\text{d}_{\mathrm{TV}}\{\pi_{\thetab}(\,\cdot\mid\bmr_h),
    \pi^{(1)}(\,\cdot\mid\bmr_h)\}
    \geq1-\delta_{\thetab}.
\end{equation*}
Since $E_h\subseteq E_{h-1}$, it follows that
\begin{equation*}
    \Pr(E_h\mid\mathcal{G}_{h-1})
    \geq(1-\delta_{\thetab})I(E_{h-1}).
\end{equation*}
Taking expectations and iterating yields $\Pr(E_H)\geq(1-\delta_{\thetab})^H$. The coupling inequality then bounds the total variation distance between the two complete rollout distributions by the probability that the coupled paths differ. Bernoulli's inequality \citep{brannan2006first} gives the final bound:
\begin{equation*}
    \text{d}_{\mathrm{TV}}\{P_{\thetab,\bmc}^{(H)},P_{1,\bmc}^{(H)}\}
    \leq\Pr(E_H^c)
    \leq1-(1-\delta_{\thetab})^H
    \leq H\delta_{\thetab}.
\end{equation*}
Thus the bound holds for every fitted kernel. Applying it to a sequence $\{\thetab_\ell\}$ with $\delta_{\thetab_\ell}\to0$ proves the asserted total variation convergence.
\end{proof}

\subsection{Bayes Risk and Amortized Prediction}\label{appdx:bayes}
The key takeaway of this subsection is that minimizing the population transformer risk over $\Thetab$ is equivalent to minimizing the average expected Kullback--Leibler divergence, $\text{d}_{\mathrm{KL}}$, from the posterior predictive rule under the training hierarchy.

Under the sampled-segment hierarchy of Section~\ref{sec:amortization}, let $\pib_{t+1}(\bmc)=(\pi_{t+1}(a\mid\bmc))_{a\in\mathcal{V}}$ denote the posterior predictive vector for a prefix $\bmc\in\mathcal{V}^t$, $t=1,\ldots,m$, where
\begin{equation*}
    \pi_{t+1}(a\mid\bmc)
    =\int \Pr_{\psib}(A_{t+1}=a\mid\mathcal{C}_t=\bmc)\,
      \Pi_{\psib}(\dif\psib\mid\mathcal{C}_t=\bmc),
    \qquad a\in\mathcal{V},
\end{equation*}
for contexts of positive probability. Because the context length identifies the local position $t$, these probabilities define a single context-based predictive rule; original trajectory locations are averaged over by the segment law. Its minimum Bayes risk is
\begin{equation*}
    \mathcal{R}^{\mathrm{Bayes}}
    =-\frac1m\sum_{t=1}^{m}
      E_{\mathcal{C}_t}\left[
        \sum_{a\in\mathcal{V}}\pi_{t+1}(a\mid\mathcal{C}_t)
        \log\{\pi_{t+1}(a\mid\mathcal{C}_t)\}
      \right].
\end{equation*}
This is the average conditional entropy given the supplied prefix, with $\psib$ unobserved. For each $\thetab\in\Thetab$, assume that $\pi_{\thetab}(a\mid\mathcal{C}_t)>0$ whenever $\pi_{t+1}(a\mid\mathcal{C}_t)>0$, almost surely, and set zero-weight terms to zero. The finite token and context spaces then imply $\mathcal{R}(\thetab)<\infty$. For probability vectors $\bmp$ and $\bmq$, the Kullback--Leibler divergence is $\text{d}_{\mathrm{KL}}(\bmp\parallel\bmq)=\sum_{a\in\mathcal{V}}p_a\log(p_a/q_a)$.

\begin{lemma}\label{lem:logrisk}
For every $\thetab\in\Thetab$,
\begin{equation*}
    \mathcal{R}(\thetab)-\mathcal{R}^{\mathrm{Bayes}}
    =\frac1m\sum_{t=1}^{m}
      E_{\mathcal{C}_t}\left[
        \text{d}_{\mathrm{KL}}\{
          \pib_{t+1}(\mathcal{C}_t)
          \parallel\pib_{\thetab}(\mathcal{C}_t)
        \}
      \right]\geq0.
\end{equation*}
Equality holds if and only if $\pib_{\thetab}(\mathcal{C}_t)=\pib_{t+1}(\mathcal{C}_t)$ almost surely for each $t=1,\ldots,m$. The posterior predictive rule uniquely minimizes the Bayes risk among unrestricted context-based predictive rules, up to contexts of probability zero.
\end{lemma}
\begin{proof}
Conditioning each summand in \eqref{eq:seqlogloss} on its prefix gives its conditional cross-entropy. Subtracting $\mathcal{R}^{\mathrm{Bayes}}$ yields the risk decomposition in Lemma~\ref{lem:logrisk}. Nonnegativity of $\text{d}_{\mathrm{KL}}$ and its equality condition establish optimality among unrestricted context-based predictive rules. No independence or stationarity of the tokens is required.
\end{proof}

For nonempty $\Thetab$, define the class approximation error,
\begin{equation*}
        \mathcal F_{\Thetab} =\inf_{\thetab\in\Thetab}\{\mathcal{R}(\thetab)-\mathcal{R}^{\mathrm{Bayes}}\}.
\end{equation*}
The predictive target is the training hierarchy's posterior rule, not each known DGP's conditional law; $\mathcal F_{\Thetab}=0$ need not imply attainment. The best population risk within the class is
\begin{equation*}
    \inf_{\thetab\in\Thetab}\mathcal{R}(\thetab)
    =\mathcal{R}^{\mathrm{Bayes}}+\mathcal F_{\Thetab}.
\end{equation*}
Thus, by Lemma~\ref{lem:logrisk}, the key result follows, i.e., minimizing the population transformer risk over $\Thetab$ is the same as minimizing the average expected $\text{d}_{\mathrm{KL}}$ from the posterior predictive rule under the training hierarchy.

\section{Monte Carlo Simulation Studies}\label{appdx:numexp}
This section evaluates: (i) recovery of context-specific next-token probabilities and (ii) generalization across orders, transition regimes, and distribution families. We use two complementary settings: a) a $p$th-order categorical process isolates the learning problem, with known next-token probabilities, and b) a vector autoregressive process that evaluates the full modeling pipeline in Fig.~\ref{fig:pipeline}, from continuous multivariate data through thresholding and tokenization. For both simulation studies we consider $d=5$.

\subsection{Simulation Settings and Details on Amortization}\label{sec:simul}
\paragraph{\textbf{a) $p$th-order categorical process.}} 
We simulate a $p$th-order categorical process, $p\geq1$, by drawing
\begin{equation*}
    A_{t+1}\mid\mathcal{C}_t=\bmc
    \sim\mathrm{Categorical}
    \bigl(\pi(0\mid\bmc),\ldots,\pi(d\mid\bmc)\bigr),
\end{equation*}
where $\bmc=(A_{t-p+1},\ldots,A_t)$.
This can be regarded as a version of \eqref{eq:next_token_distribution}
with context length $C=p$, but where the next-token probabilities are specified by an exponential tilt of a fixed baseline:
\begin{equation*}
    \pi(a\mid\bmc)
    =
    \frac{\pi_{\mathrm{base},a}
    \exp\left\{\beta\sum_{l=1}^{p}\lambda_l
    I(c_{p+1-l}=a)\right\}}
    {\sum_{b\in\mathcal{V}}\pi_{\mathrm{base},b}
    \exp\left\{\beta\sum_{l=1}^{p}\lambda_l
    I(c_{p+1-l}=b)\right\}},
    \qquad a\in\mathcal{V}.
\end{equation*}
We fix $\pib_{\mathrm{base}}=(\pi_{\text{base}, 0}, \dots, \pi_{\text{base}, d}) = (0.90,0.02,0.02,0.02,0.02,0.02)^{\T}$
and use normalized exponential lag weights, motivated by high-order mixture transition models \citep{raftery1985model},
\begin{equation*}
    \lambda_l
    =
    \frac{\exp\{-\delta(l-1)\}}
    {\sum_{r=1}^{p}\exp\{-\delta(r-1)\}},
    \qquad l=1,\ldots,p.
\end{equation*}
Here, $\delta>0$ controls the decay of influence with lag, while $\beta\geq0$ governs the strength of temporal dependence. At $\beta=0$, the next token follows the baseline distribution independently of the history; $\beta>0$ favours tokens with larger lag-weighted counts. The construction treats all active-token labels symmetrically. 

For amortized training, we sample categorical DGPs from the training prior $\Pi_{\psib}$ described in Section~\ref{sec:amortization}, with $\psib=(p,\beta,\delta)$ distributed as follows:
\begin{equation*}
    \begin{aligned}
        p&\sim\mathrm{Geom}(0.2)\mid p\in\{1,\ldots,10\}, \qquad 
        \beta&\sim\mathrm{Unif}(0,0.6), \qquad 
        \delta&\sim\mathrm{Unif}(0.05,0.55).
    \end{aligned}
\end{equation*}

Finally, we emphasize that the categorical kernel is available analytically for every simulated history, providing an exact benchmark for the transformer's next-token probabilities without Monte Carlo error in the reference values.

\paragraph{\textbf{b) Heavy-tailed $p$th order vector autoregressive process.}} 
We simulate a heavy-tailed VAR($p$) process \citep{mikosch2024extreme},
\begin{equation*}
    \bmY_t=\sum_{l=1}^{p}\bmB_l\bmY_{t-l}+\varepsilonb_t,
\end{equation*}
with coefficient matrices $\bmB_l\in\mathbb{R}^{d\times d}$ and
time-varying innovations
\begin{equation*}
    \varepsilonb_t
    =\mathrm{diag}\left(\sigma_1e^{g_{t,1}},\ldots,
    \sigma_de^{g_{t,d}}\right)\etab_t,
    \qquad
    \etab_t\sim t_{\nu}(\bm0,\Sigmab).
\end{equation*}
Here, $t_{\nu}$ is the $d$-dimensional Student-$t$ distribution ($d=5$) with $\nu>0$ degrees of freedom and positive definite scale matrix $\Sigmab$ with unit diagonal. The baseline scales $\sigmab$ are modulated by the log-volatility process $\bmg_t$, which follows a component-specific AR(1) model \citep{jacquier2004bayesian,chiu2017forecasting}:
\begin{equation*}
    \bmg_t
    =\mathrm{diag}(\varphib)\bmg_{t-1}
    +\tau_cc_t\bm1
    +\mathrm{diag}(\tau_1,\ldots,\tau_d)\xib_t.
\end{equation*}
Here, $\varphib$ contains the persistence coefficients, while $c_t$ and $\xib_t$ represent common and component-specific shocks, scaled by $\tau_c$ and $\taub$, respectively. These dynamics combine common and component-specific volatility fluctuations; setting $\bmg_t\equiv\bm0$ yields fixed innovation scales.

Using again the normalized exponential lag weights defined for the categorical process, we set
\begin{equation*}
    \bmB_l=\lambda_l\widetilde{\bmB}_l,
    \qquad l=1,\ldots,p,
\end{equation*}
where $\widetilde{\bmB}_l$ are the unnormalized coefficient matrices.
The simulated VAR process $\{\bmY_t\}_{t = 1}^n$ is then mapped to the extremal
vocabulary $\mathcal{V}$ using \eqref{eq:token}.

For amortized training, we draw the VAR parameters at each training
refresh from the prior $\Pi_{\psib}$ in Section~\ref{sec:amortization}:
\begin{equation*}
    \begin{aligned}
        p&\sim\mathrm{Geom}(0.2)\mid p\in\{1,\ldots,10\},\\
        (\widetilde{\bmB}_l)_{i,j}
        &\overset{\mathrm{iid}}{\sim}\mathrm{N}(0,1),\\
        \delta&\sim\mathrm{Unif}(0.05,0.35),\\
        \nu&\sim\mathrm{DU}(2,5),\\
        \Sigmab&\sim\mathrm{LKJ}_5(2),\\
        \sigma_j&\overset{\mathrm{iid}}{\sim}
        \mathrm{N}^{+}(0,0.15^2),\\
        \varphi_j&\overset{\mathrm{iid}}{\sim}
        \mathrm{Unif}(0.97,0.995),\\
        \tau_c,\tau_j&\overset{\mathrm{iid}}{\sim}
        \mathrm{Unif}(0.005,0.015),
    \end{aligned}
\end{equation*}
for $l=1,\ldots,p$ and $i,j=1,\ldots,d$. Here, $\mathrm{N}^{+}$ denotes a normal distribution truncated to positive values, $\mathrm{DU}$ denotes the discrete uniform distribution, and $\mathrm{LKJ}$ denotes the Lewandowski--Kurowicka--Joe correlation-matrix distribution \citep{lewandowski2009generating}. The $\mathrm{LKJ}_5(2)$ prior has the identity matrix as its mode. Within each simulated trajectory, the volatility shocks are drawn as
\begin{equation*}
    c_t,\xi_{t,j}\overset{\mathrm{iid}}{\sim}\mathrm{N}(0,1),
    \qquad t=1,\ldots,n,\quad j=1,\ldots,d.
\end{equation*}
The numerical prior settings are arbitrary, and for numerical stability, each log-volatility update is clipped componentwise to $[-1.25,1.25]$ before computing the innovation scales.

For each sampled DGP, we estimate the component-specific thresholds $u_j$ at $q=0.98$ from a separate calibration trajectory. This trajectory is generated independently of the trajectory to be tokenized, using the same DGP parameters. 

Unlike the categorical DGP, the tokenized VAR process does not provide conditional next-token probabilities in closed form. Their evaluation requires integration over latent continuous and volatility states compatible with the observed token history. We approximate these probabilities using sequential Monte Carlo particle filtering \citep{doucet2001introduction}; the \hyperref[par:particle_filter]{final paragraph} of this appendix gives implementation details and assesses the performance of our approach. 

\paragraph{\textbf{Evaluation metrics}.~}A key target of the simulation study is the conditional distribution of the next token given the cascading sequences. For a realized context of length $C$, we compare the fitted probability vector $\widehat{\pib}$ with the DGP's reference vector $\pib$. We consider the Kullback--Leibler divergence and the total variation distance,
\begin{equation*}
    \text{d}_{\mathrm{KL}}\left(\pib,\ \widehat{\pib}\right)=\sum_{j=0}^{\lvert\mathcal{V}\rvert-1}\pi_{j}\log \left(\frac{{\pi}_{j}}{\widehat{\pi}_j}\right), \qquad 
    \text{d}_{\mathrm{TV}} \left(\pib, \ \widehat{\pib}\right)= \frac{1}{2}\sum_{j=0}^{\lvert\mathcal{V}\rvert-1}\left\lvert\pi_{j}- \widehat{\pi}_j\right\rvert.
\end{equation*}
All discrepancy measures are computed first at the individual history level and then aggregated within each DGP and order by taking the mean. Finally, we report $\lvert\pi_0-\widehat{\pi}_0\rvert$, the absolute error in the conditional probability of a non-quiet token.

\paragraph{\textbf{Model configuration}}
The training, validation, and testing DGPs are generated independently but they share the same conditional DGP families and hyperparameter priors specified above. The training set samples the different orders from the truncated geometric prior described above, whereas the test sets are balanced across $p$ to permit comparable order-specific evaluation. Hence the test sets represent unseen realizations of the same amortization family. 

The training data are organised into $M_{\text{train}}=10\,000$ training refreshes, with $32$ different regimes generated within each refresh; each regime uses a new set of parameters and autoregressive order $p\in\{1,\ldots,10\}$, generated from the prior. From each refresh, $4,096$ segments of length $C+1 = 257$ are sampled, with the temporal ordering preserved within every segment. The trajectories are then represented as a sequence of extremal tokens via the probability-scale rule defined in \eqref{eq:token}. This results in $1\,048\,576$ target positions per epoch during training. Each refresh is used for two consecutive epochs before the trajectory is refreshed. No early stopping rule is imposed during training. For evaluation, we construct a balanced test set by generating $25$ sets of test parameters for each order $p$, giving $M_{\text{test}}=250$ unseen test datasets.

Training uses a decoder-only transformer with $L\!=\!4$ layers, $H\!=\!8$ attention heads, $d_{\text{emb}}\!=\!128$, feed forward dimension $512$ with Gaussian Error Linear Unit \citep[GELU, ][]{hendrycks2016gaussian} activation  and dropout \citep{srivastava2014dropout} rate $0.1$. This compact specification is fixed across the amortization and is chosen to provide sufficient capacity while keeping the simulation computationally manageable. Also, the transformer does not receive as input the order, thresholding quantile, transition effects, lag decay parameter, and true transition parameters.

Thus, our model architecture contains $827\,654$ learnable parameters.
For optimization, we use a stochastic gradient descent based algorithm, AdamW \citep{loshchilovH19decoupled} with batch size of $32$ to minimize the cross entropy loss at all shifted target positions. Specific details on the optimization routine are described in \appendixname~\ref{apdx:specs}.

\subsection{Results from Simulation Studies}
\paragraph{\textbf{a) $p$th-order categorical process.}} 
\tablename~\ref{tab:condrecov} (top) first reports the results for the categorical process, for which the conditional probabilities are available in closed form. The transformer recovers these probabilities closely across all orders. Mean $\text{d}_{\mathrm{TV}}$ ranges from $0.0079$ to $0.0105$, while mean $\text{d}_{\mathrm{KL}}$ ranges from $0.0007$ to $0.0012$. The largest mean discrepancy occurs at $p=2$, with $\text{d}_{\mathrm{TV}}$ of $0.0105$ and $\text{d}_{\mathrm{KL}}$ of $0.0012$, but the variation across orders is small and there is no indication of deteriorating recovery as $p$ increases. The conditional probability vectors are likewise closely matched. The exact active mass, $1-\pi_0$ ranges from $0.066$ to $0.077$, compared with transformer estimates between $0.070$ and $0.075$. Even at $p=8$, where the difference in active mass is largest, the exact value is $0.066$ compared with $0.071$ estimated by the transformer. For several orders, including $p=4$, $p=6$, and $p=7$, the two active masses agree to three decimal places. The history-wise $\text{d}_{\mathrm{KL}}$ and $\text{d}_{\mathrm{TV}}$ therefore provide direct evidence that the transformer recovers the conditional distributions accurately, while the averaged probability vectors show that the remaining local discrepancies do not induce a systematic distortion in the allocation of probability between quiet and active tokens.

\paragraph{\textbf{b) Heavy-tailed $p$th order vector autoregressive process.}} 
\tablename~\ref{tab:condrecov} reports the simulation results for the heavy-tailed VAR($p$) process, for which the reference conditional probabilities are approximated using Monte Carlo. Agreement remains close across orders $p=1,\ldots,10$, where mean $\text{d}_{\mathrm{KL}}$ ranges from $0.0056$ to $0.0128$, while mean $\text{d}_{\mathrm{TV}}$ lies between $0.0161$ and $0.0237$. The largest mean discrepancy occurs at $p=1$, with $\text{d}_{\mathrm{KL}}=0.0128$ and $\text{d}_{\mathrm{TV}}=0.0237$, whereas the remaining orders exhibit comparably small errors. Importantly, there is no systematic deterioration in conditional probability recovery as the dependence order increases. The history-averaged active mass is also reproduced closely: the Monte Carlo reference values range from $0.069$ to $0.082$, compared with transformer estimates between $0.071$ and $0.084$. Averaged across orders, the Monte Carlo and transformer active masses are approximately $0.0766$ and $0.0778$, respectively. Thus, despite the additional difficulty introduced by thresholding a continuous heavy-tailed process and approximating its conditional token probabilities using Monte Carlo, the amortized transformer preserves the overall allocation between quiet and active tokens and maintains stable conditional recovery across the full range of orders considered.

\begin{table}[!htbp]
\centering
\caption{\footnotesize Conditional probability recovery on the test set for categorical process (top subtable) and vector autoregressive process (bottom subtable). For each order $p$, $\pib$ and $\widehat{\pib}$ are the mean conditional probability vectors across the test set. 
Here, $\text{d}_{\mathrm{KL}}$ and $\text{d}_{\mathrm{TV}}$ are computed for each history and then averaged within each order. The final two columns report the corresponding averaged active event probabilities.}\label{tab:condrecov}
\small
\resizebox{\textwidth}{!}{
\begin{tabular}{
>{\centering\arraybackslash}p{0.45cm}
>{\centering\arraybackslash}p{4.60cm}
>{\centering\arraybackslash}p{4.60cm}
>{\centering\arraybackslash}p{1.60cm}
>{\centering\arraybackslash}p{1.60cm}
>{\centering\arraybackslash}p{1.40cm}
>{\centering\arraybackslash}p{1.40cm}
}
\toprule
$p$
& $\pib$
& $\widehat{\pib}$
& $\overline{d}_{\mathrm{KL}}(\pib,\widehat{\pib})$
& $\overline{d}_{\mathrm{TV}}(\pib,\widehat{\pib})$
& $1-\pi_0$
& $1-\widehat{\pi}_0$ \\
\midrule

1
& $(0.929,0.014,0.015,0.014,0.014,0.014)$
& $(0.928,0.014,0.015,0.015,0.014,0.014)$
& $0.0009$ & $0.0089$ & $0.071$ & $0.072$ \\

2
& $(0.928,0.014,0.014,0.014,0.014,0.014)$
& $(0.928,0.014,0.014,0.015,0.014,0.014)$
& $0.0012$ & $0.0105$ & $0.072$ & $0.072$ \\

3
& $(0.924,0.015,0.015,0.015,0.015,0.015)$
& $(0.927,0.015,0.015,0.015,0.015,0.015)$
& $0.0008$ & $0.0083$ & $0.076$ & $0.073$ \\

4
& $(0.927,0.015,0.015,0.015,0.015,0.015)$
& $(0.927,0.014,0.015,0.015,0.015,0.015)$
& $0.0007$ & $0.0084$ & $0.073$ & $0.073$ \\

5
& $(0.932,0.014,0.014,0.014,0.014,0.014)$
& $(0.930,0.014,0.014,0.014,0.014,0.014)$
& $0.0010$ & $0.0091$ & $0.068$ & $0.070$ \\

6
& $(0.929,0.014,0.014,0.014,0.014,0.014)$
& $(0.929,0.014,0.014,0.014,0.014,0.014)$
& $0.0009$ & $0.0092$ & $0.071$ & $0.071$ \\

7
& $(0.925,0.015,0.015,0.015,0.015,0.015)$
& $(0.925,0.015,0.015,0.015,0.015,0.015)$
& $0.0007$ & $0.0081$ & $0.075$ & $0.075$ \\

8
& $(0.934,0.013,0.013,0.013,0.013,0.013)$
& $(0.929,0.014,0.014,0.014,0.014,0.014)$
& $0.0008$ & $0.0082$ & $0.066$ & $0.071$ \\

9
& $(0.923,0.015,0.015,0.015,0.015,0.015)$
& $(0.925,0.015,0.015,0.015,0.015,0.015)$
& $0.0008$ & $0.0079$ & $0.077$ & $0.075$ \\

10
& $(0.929,0.014,0.014,0.014,0.014,0.014)$
& $(0.928,0.014,0.014,0.015,0.015,0.014)$
& $0.0011$ & $0.0096$ & $0.071$ & $0.072$ \\

\bottomrule
\end{tabular}}
\resizebox{\textwidth}{!}{
\begin{tabular}{
>{\centering\arraybackslash}p{0.45cm}
>{\centering\arraybackslash}p{4.60cm}
>{\centering\arraybackslash}p{4.60cm}
>{\centering\arraybackslash}p{1.60cm}
>{\centering\arraybackslash}p{1.60cm}
>{\centering\arraybackslash}p{1.40cm}
>{\centering\arraybackslash}p{1.40cm}
}
$p$
& $\overline{\pib}^{\mathrm{MC}}$
& $\widehat{\pib}$
& $\overline{d}_{\mathrm{KL}}(\overline{\pib}^{\mathrm{MC}}, \widehat{\pib})$
& $\overline{d}_{\mathrm{TV}}(\overline{\pib}^{\mathrm{MC}}, \widehat{\pib})$
& $1-\overline{\pib}^{\mathrm{MC}}_0$
& $1-\widehat{\pi}_0$ \\
\midrule

1
& $(0.918,0.016,0.019,0.014,0.018,0.015)$
& $(0.917,0.017,0.017,0.016,0.017,0.016)$
& $0.0128$ & $0.0237$ & $0.082$ & $0.083$ \\

2
& $(0.922,0.013,0.016,0.019,0.014,0.015)$
& $(0.922,0.015,0.015,0.017,0.015,0.015)$
& $0.0067$ & $0.0191$ & $0.078$ & $0.078$ \\

3
& $(0.921,0.016,0.017,0.017,0.015,0.014)$
& $(0.920,0.017,0.016,0.016,0.016,0.015)$
& $0.0079$ & $0.0202$ & $0.079$ & $0.080$ \\

4
& $(0.927,0.015,0.014,0.014,0.016,0.014)$
& $(0.925,0.016,0.014,0.015,0.016,0.014)$
& $0.0064$ & $0.0181$ & $0.073$ & $0.075$ \\

5
& $(0.922,0.016,0.015,0.013,0.015,0.018)$
& $(0.923,0.016,0.014,0.014,0.016,0.017)$
& $0.0067$ & $0.0178$ & $0.078$ & $0.077$ \\

6
& $(0.931,0.014,0.012,0.014,0.013,0.015)$
& $(0.929,0.015,0.012,0.015,0.014,0.015)$
& $0.0056$ & $0.0172$ & $0.069$ & $0.071$ \\

7
& $(0.931,0.014,0.012,0.014,0.015,0.013)$
& $(0.929,0.015,0.014,0.014,0.015,0.014)$
& $0.0058$ & $0.0161$ & $0.069$ & $0.071$ \\

8
& $(0.919,0.018,0.019,0.014,0.015,0.015)$
& $(0.916,0.018,0.018,0.016,0.016,0.017)$
& $0.0060$ & $0.0178$ & $0.081$ & $0.084$ \\

9
& $(0.922,0.016,0.016,0.015,0.017,0.014)$
& $(0.920,0.017,0.016,0.016,0.017,0.014)$
& $0.0059$ & $0.0173$ & $0.078$ & $0.080$ \\

10
& $(0.921,0.012,0.016,0.017,0.015,0.019)$
& $(0.921,0.014,0.016,0.016,0.016,0.018)$
& $0.0062$ & $0.0184$ & $0.079$ & $0.079$ \\
\bottomrule
\end{tabular}}
\end{table}

\begin{figure}[!htbp]
    \centering
    \includegraphics[width=0.97\linewidth]{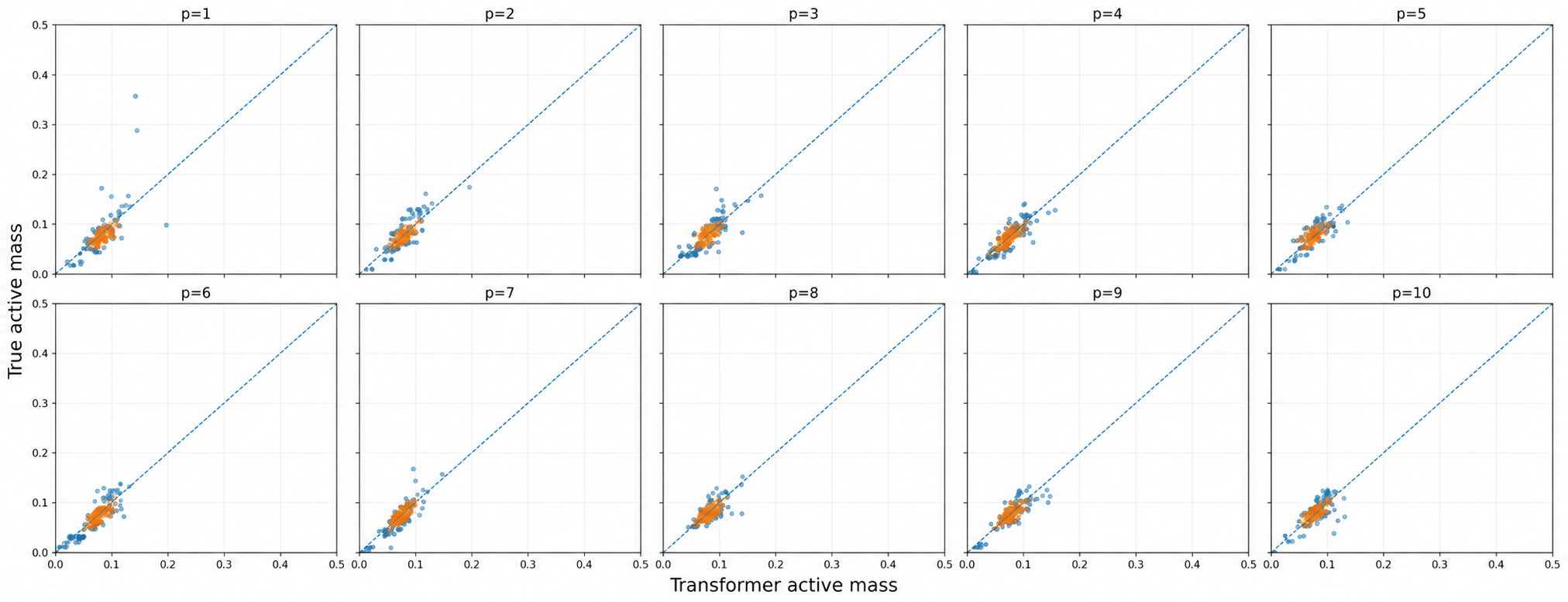}
    \caption{\footnotesize Scatter plots of active mass $1-\pi_0$ against $1-\widehat{\pi}_0$ for categorical DGP (orange) and VAR DGP (blue), by true order $p$. The diagonal line marks exact calibration.}
    \label{fig:activemass}
\end{figure}

\paragraph{\textbf{Other takeaways from numerical experiments.}} 
Fig.~\ref{fig:activemass} reinforces the results in \tablename~\ref{tab:condrecov}, showing close agreement between estimated and reference active-event probabilities for both DGP families. Across most test histories, the estimated and reference probabilities lie close to the $45$ degree line, indicating that the transformer recovers not only the allocation of probability across individual tokens but also the aggregate probability assigned to an active event. The categorical process exhibits particularly close agreement: its mean $\text{d}_{\mathrm{TV}}$ ranges from only $0.0079$ to $0.0105$ across $p=1,\ldots,10$, and the context averaged active masses differ by at most approximately $0.005$. The VAR process displays somewhat greater contextwise variability, consistent with its larger mean $\text{d}_{\mathrm{TV}}$ values of $0.0161$--$0.0237$. The largest mean discrepancy again occurs at $p=1$, but the errors remain small and there is no systematic deterioration as the dependence order increases. The VAR active masses in \tablename~\ref{tab:condrecov} show similarly close agreement, with Monte Carlo reference values between $0.069$ and $0.082$ and transformer estimates between $0.071$ and $0.084$. Moreover, the difference in active mass changes sign across orders rather than exhibiting a persistent directional bias. Overall, Fig.~\ref{fig:activemass} complements the averaged $\text{d}_{\mathrm{KL}}$ and $\text{d}_{\mathrm{TV}}$ summaries in \tablename~\ref{tab:condrecov} by showing that the small average discrepancies are not solely an artefact of aggregation across histories; for the majority of individual test histories, the conditional probability assigned to the occurrence of an active event is recovered closely.

\begin{figure}[!htbp]
    \centering
    \includegraphics[width=0.99\linewidth]{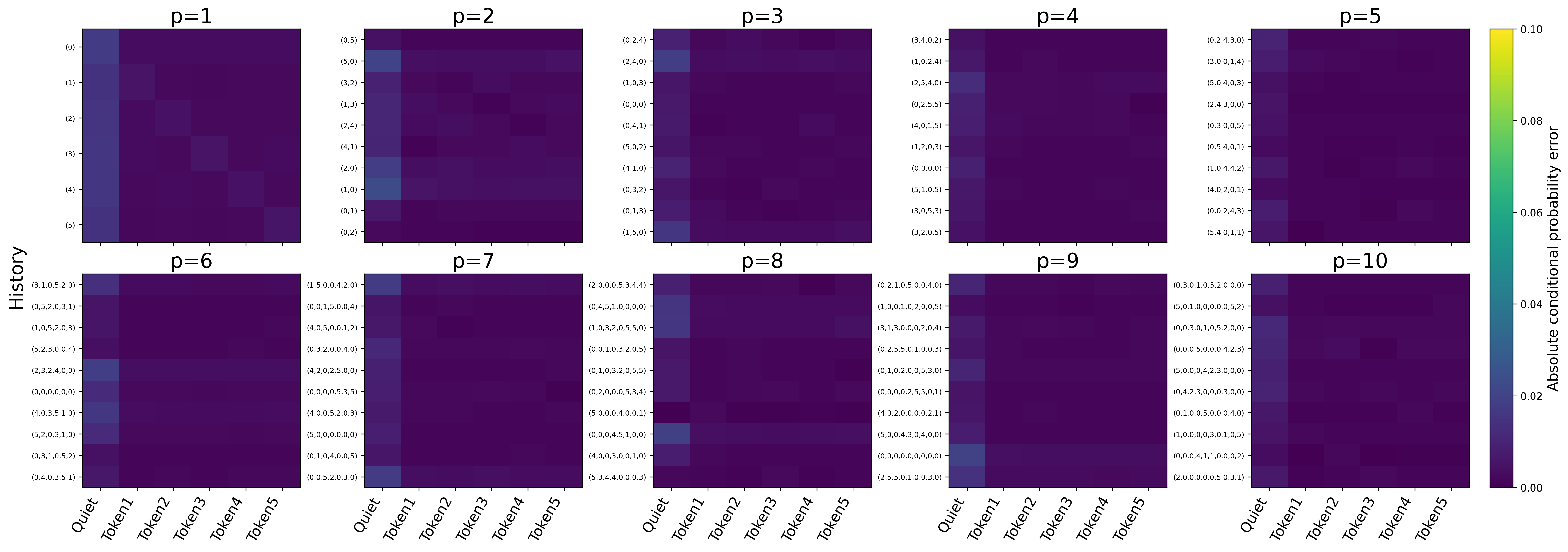}
    \caption{\footnotesize Within DGP conditional probability error heatmaps for categorical DGP. Each plot corresponds to testing DGP of true order $p$; rows are distinct histories and columns are the subsequent values in the extremal vocabulary. Each tile shows the absolute conditional probability error and each tick on the left hand side is the testing history.}
    \label{fig:hisperp}
\end{figure}

Fig.~\ref{fig:hisperp} shows that the transformer recovers the exact next token distribution accurately across a broad range of contexts and dependence orders. Most cells are dark, indicating small token-wise absolute probability errors. Averaged across context, the conditional MAE ranges from $0.0022$ to $0.0052$, while the corresponding mean $\text{d}_{\mathrm{TV}}$ ranges from $0.0066$ to $0.0155$ across $p=1,\ldots,10$. The smallest mean errors occur at $p=5$, where the MAE and $\text{d}_{\mathrm{TV}}$ are $0.0022$ and $0.0066$, respectively, while even at the largest mean discrepancy, occurring at $p=1$, these remain only $0.0052$ and $0.0155$. The largest cell-wise discrepancies are concentrated in the \textrm{Quiet} token, whose mean absolute error across all displayed histories is approximately $0.0092$, compared with roughly $0.0020$--$0.0022$ for each active token. 

\begin{table}[!htbp]
\centering
\scriptsize
\caption{\footnotesize Context specific recovery of the fixed $10$th-order conditional distribution. For each context $\mathcal{C}_t$, $A_{t+1}$ is the realized next token and the vectors contain the tokenwise absolute conditional probability errors $|\widehat{\pi}_j-\pi^{\mathrm{MC}}_{j}|$ with conditioning length $\ell=1, 5,$ and $10$ for the corresponding Fig.~\ref{fig:fixedhis}.}\label{tab:pmax}
\resizebox{\textwidth}{!}{
\begin{tabular}{ccccc}
\toprule
$\mathcal{C}_t$ & $A_{t+1}$ & $\ell=1$ & $\ell=5$ & $\ell=10$ \\
\midrule
$(0,0,0,2,2,0,0,0,4,2)$ & 2 &
$(0.0353,0.0119,0.0041,0.0113,0.0029,0.0134)$ &
$(0.0575,0.0157,0.0057,0.0157,0.0145,0.0173)$ &
$(0.0570,0.0132,0.0059,0.0123,0.0105,0.0151)$ \\

$(0,0,0,0,3,0,0,2,3,3)$ & 2 &
$(0.0264,0.0174,0.0294,0.0142,0.0131,0.0111)$ &
$(0.0670,0.0250,0.0117,0.0164,0.0200,0.0172)$ &
$(0.0595,0.0223,0.0155,0.0195,0.0178,0.0154)$ \\

$(3,0,0,0,0,2,0,0,1,3)$ & 3 &
$(0.0276,0.0116,0.0148,0.0051,0.0137,0.0056)$ &
$(0.0070,0.0245,0.0032,0.0037,0.0094,0.0012)$ &
$(0.0037,0.0215,0.0039,0.0018,0.0099,0.0021)$ \\

$(0,0,0,0,3,2,0,1,4,2)$ & 2 &
$(0.0720,0.0134,0.0300,0.0320,0.0150,0.0116)$ &
$(0.0035,0.0041,0.0189,0.0208,0.0329,0.0008)$ &
$(0.0154,0.0012,0.0258,0.0177,0.0301,0.0032)$ \\

$(0,0,0,0,0,4,0,1,5,4)$ & 5 &
$(0.0362,0.0239,0.0114,0.0063,0.0161,0.0106)$ &
$(0.0125,0.0098,0.0032,0.0011,0.0217,0.0028)$ &
$(0.0056,0.0111,0.0042,0.0000,0.0181,0.0028)$ \\

$(1,0,0,0,0,2,0,0,3,3)$ & 2 &
$(0.0104,0.0087,0.0195,0.0128,0.0043,0.0087)$ &
$(0.0132,0.0120,0.0119,0.0042,0.0065,0.0109)$ &
$(0.0181,0.0178,0.0104,0.0059,0.0069,0.0099)$ \\

$(0,0,0,3,5,0,0,0,5,5)$ & 2 &
$(0.0200,0.0000,0.0155,0.0009,0.0038,0.0325)$ &
$(0.0132,0.0009,0.0160,0.0030,0.0025,0.0306)$ &
$(0.0002,0.0011,0.0169,0.0094,0.0017,0.0254)$ \\

$(5,0,5,0,0,3,0,0,4,4)$ & 3 &
$(0.0178,0.0199,0.0066,0.0125,0.0008,0.0309)$ &
$(0.0004,0.0212,0.0093,0.0062,0.0043,0.0290)$ &
$(0.0117,0.0212,0.0079,0.0041,0.0048,0.0181)$ \\

$(0,0,3,0,0,0,0,5,5,2)$ & 4 &
$(0.0121,0.0108,0.0175,0.0075,0.0024,0.0105)$ &
$(0.0289,0.0176,0.0178,0.0148,0.0058,0.0085)$ &
$(0.0271,0.0163,0.0199,0.0194,0.0046,0.0068)$ \\

$(0,0,0,5,0,0,0,5,4,3)$ & 5 &
$(0.0567,0.0000,0.0070,0.0143,0.0123,0.0903)$ &
$(0.0148,0.0082,0.0154,0.0140,0.0250,0.0773)$ &
$(0.0182,0.0061,0.0140,0.0130,0.0224,0.0737)$ \\
\bottomrule
\end{tabular}}
\end{table}

\begin{figure}[!htbp]
    \centering
    \includegraphics[width=0.99\linewidth]{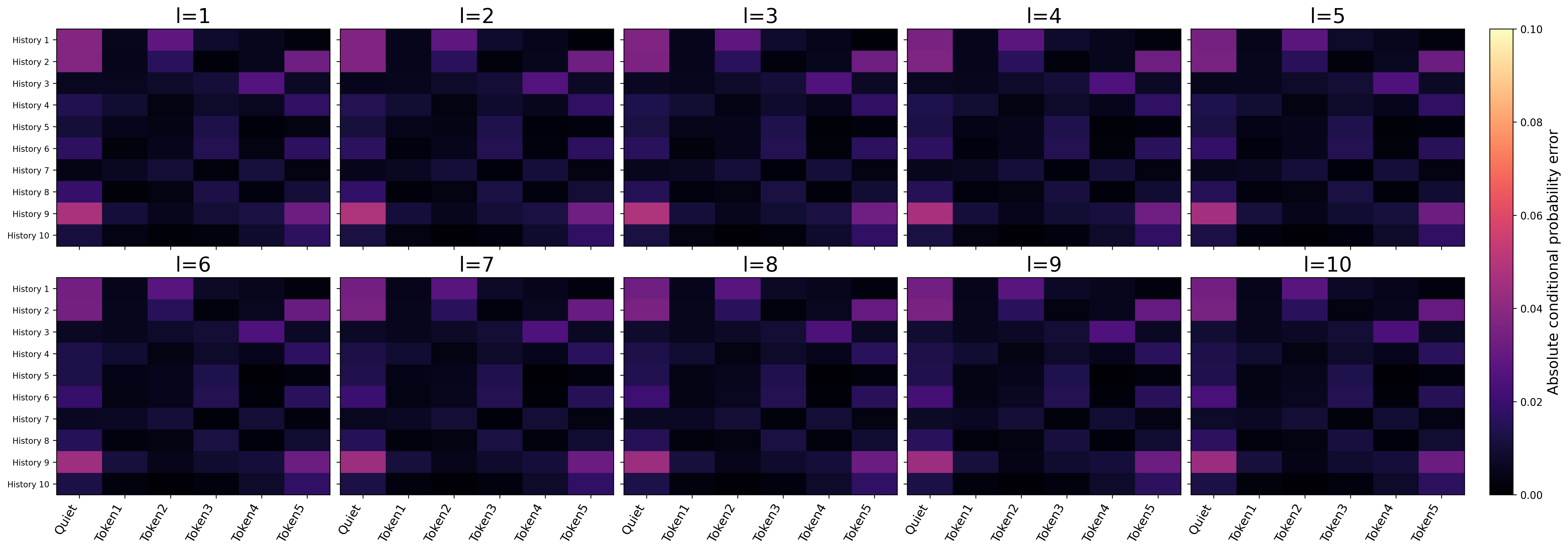}
    \caption{\footnotesize Within-DGP fixed-target diagnostic for the $p\!=\!10$ VAR DGP. Each row represents one of the selected contexts in \tablename~\ref{tab:pmax}, each column represents an extremal token, and each plot corresponds to a context length supplied to the transformer. The color scale is absolute conditional probability error.}
    \label{fig:fixedhis}
\end{figure}

Lastly, \tablename~\ref{tab:pmax} and Fig.~\ref{fig:fixedhis} examine how well the transformer approximates the fixed $10$th-order conditional distribution when only a truncated suffix of the history is observed. Across the selected cascade-rich histories, recovery is already relatively accurate for short contexts and improves as additional lag information becomes available. Averaged across histories and tokens, the mean absolute probability error decreases from $0.0169$ at $\ell=1$ to $0.0146$ at $\ell=10$, while the corresponding mean total variation decreases from $0.0507$ to $0.0438$. The largest discrepancies are concentrated in the Quiet-token probability, whose mean absolute error decreases from $0.0314$ to $0.0217$, whereas the mean error across the individual active tokens is substantially smaller and decreases from $0.0140$ to $0.0132$. Thus, the aggregate results indicate that the amortized transformer makes effective use of additional lag information encoded by the VAR process while retaining accurate conditional probability estimates under truncated histories.

The simulations support the transformer as an amortized estimator of conditional token probabilities across the transition mechanisms and context lengths considered.

\phantomsection
\paragraph{\textbf{Particle filter reference probabilities.}}\label{par:particle_filter}
For each heavy-tailed VAR DGP and token history, we approximate next-token probabilities by particle filtering \citep{doucet2001introduction}. We initialize $8192$ particles after a $1000$-step burn-in, condition on successive observed tokens using binary matching weights, and resample. Fewer than eight matching particles trigger a complete restart with twice as many particles, up to $131\,072$. In a complete repeat $b$, we draw $N_b$ next tokens from the particle predictive mixture. For $B=3$ independent filtering-and-prediction repeats, define
\begin{equation*}
    \widehat{\pi}^{\mathrm{MC},(b)}_{j}
    =\frac{1}{N_b}\sum_{r=1}^{N_b}I(A_{t+1}^{(b,r)}=j),
    \qquad
    \overline{\pi}^{\mathrm{MC}}_{j}
    =\frac{1}{B}\sum_{b=1}^{B}\widehat{\pi}^{\mathrm{MC},(b)}_{j}.
\end{equation*}
The Monte Carlo standard error MCSE of the average is estimated by
\begin{equation*}
    \mathrm{MCSE}_j
    =\left\{\frac{1}{B(B-1)}
      \sum_{b=1}^{B}
      (\widehat{\pi}^{\mathrm{MC},(b)}_j-\overline{\pi}^{\mathrm{MC}}_j)^2
      \right\}^{1/2}.
\end{equation*}
Complete-repeat variation includes both filtering and predictive sampling uncertainty; this MCSE does not assess systematic approximation error arising from the finite particle population or finite burn-in. Table~\ref{tab:mcse} summarizes Monte Carlo uncertainty by process order $p$. The particle filter reference probabilities exhibit low Monte Carlo uncertainty across all latent VAR orders, with median absolute MCSEs remaining around $3\times10^{-4}$ and median relative MCSEs below approximately $2\%$. Even in the upper tail of the distribution, the 95th percentile relative MCSE remains below approximately $8.2\%$, indicating that Monte Carlo variability remains well controlled across process orders.


\begin{table}[!htbp]
\centering
\caption{\footnotesize MCSE of the particle filter approximation of next-token probabilities. The median and 95th percentile are computed across all DGP, context and token probabilities for each order, and relative MCSE is expressed as a percentage in parentheses. The MCSE is computed from $B=3$ independent complete filtering-and-prediction repeats as the between-repeat sample standard deviation divided by $\sqrt{B}$.}\label{tab:mcse}

\resizebox{\textwidth}{!}{
\begin{tabular}{cccccc}
\toprule
$p$
& $1$ & $2$ & $3$ & $4$ & $5$ \\
\midrule
Median $\mathrm{MCSE}$
& $0.000333$ ($1.90$)
& $0.000323$ ($1.81$)
& $0.000332$ ($2.01$)
& $0.000284$ ($1.87$)
& $0.000329$ ($1.98$) \\

95th percentile $\mathrm{MCSE}$
& $0.001682$ ($8.17$)
& $0.001511$ ($7.55$)
& $0.001598$ ($7.13$)
& $0.001463$ ($6.59$)
& $0.001519$ ($7.61$) \\

\midrule
$p$
& $6$ & $7$ & $8$ & $9$ & $10$ \\
\midrule

Median $\mathrm{MCSE}$
& $0.000275$ ($1.92$)
& $0.000287$ ($1.88$)
& $0.000327$ ($1.81$)
& $0.000356$ ($1.90$)
& $0.000313$ ($1.92$) \\

95th percentile $\mathrm{MCSE}$
& $0.001380$ ($6.61$)
& $0.001352$ ($6.90$)
& $0.001291$ ($6.72$)
& $0.001457$ ($6.75$)
& $0.001433$ ($6.96$) \\

\bottomrule
\end{tabular}}
\end{table}

\phantomsection
\paragraph{\textbf{Robustness of the probability-scale rule}}
Entries below report Monte Carlo medians, with the $97.5\%$ credible interval across 500 replications.

\begin{table}[!htbp]
\centering
\caption{\footnotesize Monte Carlo robustness of the probability-scale marginal transformation and the induced extremal token structure.}
\label{tab:mc_fixed_empirical_robustness}
\small

\textbf{Panel A: Fixed-empirical transformation evaluation}

\vspace{0.10cm}

\begin{tabular}{lcc}
\toprule
Metric & Target & Fixed empirical \\
\midrule
Mean PIT
& $0.5000$
& 0.4999 [0.4985, 0.5013] \\

PIT variance
& $0.08333$
& 0.08333 [0.08268, 0.08398] \\

Full-distribution QQ RMSE
& $0$
& 0.00182 [0.00122, 0.00265] \\

Upper-tail probability, $\Pr(U>0.98)$
& $0.0200$
& 0.01997 [0.01908, 0.02092] \\

Upper-tail QQ RMSE
& $0$
& 0.000232 [0.000168, 0.000353] \\
\bottomrule
\end{tabular}

\vspace{0.35cm}

\textbf{Panel B: Agreement with the fixed-empirical benchmark}

\vspace{0.10cm}

\resizebox{\textwidth}{!}{
\begin{tabular}{lccc}
\toprule
Metric
& Fixed GPD
& Time-varying GPD
& Time-varying Harvey \\
\midrule

Active-region CDF MAE
& 0.00010 [0.00008, 0.00014]
& 0.00019 [0.00013, 0.00029]
& 0.0106 [0.00933, 0.0123] \\

\addlinespace[2pt]

Exceedance-indicator agreement (\%)
& 100.0 [100.0, 100.0]
& 100.0 [100.0, 100.0]
& 100.0 [100.0, 100.0] \\

Extreme-token agreement (\%)
& 99.72 [99.61, 99.80]
& 99.58 [99.40, 99.71]
& 95.61 [95.09, 96.02] \\

Cascade-token agreement (\%)
& 99.51 [99.30, 99.69]
& 99.28 [98.97, 99.55]
& 92.47 [91.48, 93.23] \\

\addlinespace[2pt]

Conditional nonzero marginal TV (\%)
& 0.046 [0.015, 0.092]
& 0.072 [0.023, 0.146]
& 0.531 [0.224, 1.019] \\

\addlinespace[2pt]

Frequency-weighted transition TV (\%)
& 0.013 [0.008, 0.019]
& 0.019 [0.011, 0.029]
& 0.109 [0.080, 0.146] \\

Extreme-source transition TV (\%)
& 0.115 [0.075, 0.173]
& 0.166 [0.096, 0.279]
& 0.983 [0.727, 1.288] \\

\bottomrule
\end{tabular}
}
\end{table}

\section{Transformer and Optimizer Specifications}\label{apdx:specs}
\thispagestyle{fancy}
\subsection{Transformer Architecture and Forward Pass}
Algorithm~\ref{alg:trans} gives the complete forward pass for the architecture in Section~\ref{subsec:architecture}, with block- and head-specific parameters; it returns next-token probability vectors at all input positions.

\begin{algorithm}[!htbp]
\caption{Decoder-only autoregressive transformer}
\label{alg:trans}
    \begin{algorithmic}[1]
    \Require 
    \Statex Sequence of extremal tokens $\{A_t\}_{t=1}^m$, with $A_t$ as in \eqref{eq:token}, and $m\leq C$
    \Statex Attention weights and biases for $l\!=\!1,\ldots,L$ and $h\!=\!1,\ldots,H$: $\bmW_h^{(\text{q},l)}, \bmW_h^{(\text{k},l)}, \bmW_h^{(\text{v},l)}$ and $\bmb_h^{(\text{q},l)},\bmb_h^{(\text{k},l)},\bmb_h^{(\text{v},l)}$
    \Statex Multi-layer perceptron parameters for $L$ layers: 
    $\bmW^{(1,l)}, \bmW^{(2,l)}$ and $\bmb^{(1,l)}, \bmb^{(2,l)}$
    \Statex Token-embedding, position-embedding, and layer-normalization parameters
    \Statex Final output parameters $\bmW^{(\mathrm{out})} \in \mathbb{R}^{d_{\text{emb}} \times |\mathcal{V}|}$ and $\bmb^{(\mathrm{out})}\in\mathbb{R}^{|\mathcal{V}|}$
    \Ensure Estimated conditional probability matrix 
    $\Pib_{\thetab} \in [0,1]^{m \times |\mathcal{V}|}$
    \Statex
    \State $\bmX^{(0)} \gets 
    \mathrm{TokenEmbed}(\{A_t\}_{t=1}^m) 
    + \mathrm{PositionalEmbed}(1{:}m)$

    \For{$l = 1, \dots, L$}

        \State $\bmX^{(l)'} 
        \gets \mathrm{LayerNorm}^{(l)}_1\!\left[\bmX^{(l-1)}\right]$
        
        \State $\bmZ^{(l)}
        \gets \bmX^{(l-1)}
        + \mathrm{MHSA}^{(l)}\!\left(\bmX^{(l)'};\bmM\right)$

        \State $\bmZ^{(l)'}
        \gets \mathrm{LayerNorm}^{(l)}_2\!\left[\bmZ^{(l)}\right]$

        \State $\bmX^{(l)}
        \gets \bmZ^{(l)}
        + \mathrm{GELU}\!\left(
        \bmZ^{(l)'}\bmW^{(1,l)}+\bm1_m(\bmb^{(1,l)})^{\T}
        \right)\bmW^{(2,l)}+\bm1_m(\bmb^{(2,l)})^{\T}$

    \EndFor

    \State $\widetilde{\bmX} 
    \gets \mathrm{LayerNorm}_{f}\!\left[\bmX^{(L)}\right]$

    \State ${\Pib}_{\thetab} 
    \gets \mathrm{Softmax}_{\mathrm{row}}\!\left(
    \widetilde{\bmX}\bmW^{(\mathrm{out})}+\bm1_m(\bmb^{(\mathrm{out})})^{\T}
    \right)$

    \State \Return ${\Pib}_{\thetab}$
    \end{algorithmic}
\end{algorithm}

\subsection{Optimization and Computing Resources}
The model is optimized using AdamW \citep{loshchilovH19decoupled} with the initial learning rate at $10^{-3}$, the decoupled weight decay coefficient is $10^{-4}$, and the first and second moment parameters are $(\beta_1,\beta_2)=(0.9,0.95)$. Optimization uses a batch size of $32$ and the Euclidean norm of the gradient is clipped at $1$ before each optimizer update. The learning rate is updated once after every epoch using cosine annealing \citep{loshchilov2017sgdr}, without warm-up or cosine restarts. For related theoretical developments on the optimization algorithms for LLMs, see \citet{liu2025muonscalablellmtraining}, \citet{lytras2026tamed} and \citet{makras2026muon}. For epoch $e\!=\!1,\ldots,20{\,}000$, the scheduler follows
\begin{equation*}
    \zeta_e = \zeta_{\min} + \frac{1}{2} \left(\zeta_0-\zeta_{\min}\right) \left\{ 1+\cos\left(\frac{\pi e}{20{,}000}\right) \right\}, \qquad \zeta_0=10^{-3},\ \zeta_{\min}=10^{-6}.
\end{equation*}
Intermediate checkpoints are saved every $1{\,}000$ epochs, and training diagnostics are logged every $25$ mini-batches and validation is performed periodically during optimization. The validation set uses a larger batch size of $4\,096$ as it does not require gradient evaluation. 

All data generation and transformer training were performed on the University of Edinburgh’s Eddie cluster (Edinburgh Compute and Data Facility), using a single NVIDIA A100 Tensor Core GPU with TensorFloat-32 (TF32) precision and 20 CPU cores on Intel Xeon Gold 6338 processors (2.00 GHz). The initial $20{\,}000$-epoch amortization training required approximately 10 hours, while the 200-epoch adaption to the semiconductor data required approximately 2 hours. Evaluations on the heavy-tailed VAR DGP required generating Monte Carlo samples via SMC particle filter and took approximately 10 minutes; for the categorical DGP with known conditional probability, evaluation took 6 minutes.

\section{Supplementary Empirical Details for Semiconductor Stocks} \label{appdx:eda}
This section presents the exploratory analysis on the negative log returns of the five semiconductor stocks. 

The dominant-exceedance representation entails a loss of contemporaneous information only when multiple marginal thresholds are exceeded simultaneously. In the ten-second intraday financial data, such events account for only $1.87\%$ of the complete timeline. Conditional on observing at least one exceedance, $75.03\%$ of observations have a unique extreme component, while $24.97\%$ have multiple exceedances. Moreover, simultaneous events are predominantly low-dimensional, with two-component exceedances accounting for $61.64\%$, and two- or three-component exceedances jointly account for $84.94\%$ of all simultaneous exceedance observations. The conditional frequency of multiple exceedances is also approximately equal between the train and test samples ($25.03\%$ and $24.36\%$, respectively), despite a higher unconditional exceedance frequency in the test period. These results support the dominant-exceedance token as a parsimonious representation of the identity of the most severe marginal extreme.

\begin{figure}[!htbp]
    \centering
    \includegraphics[width=0.97\linewidth, height=4cm]{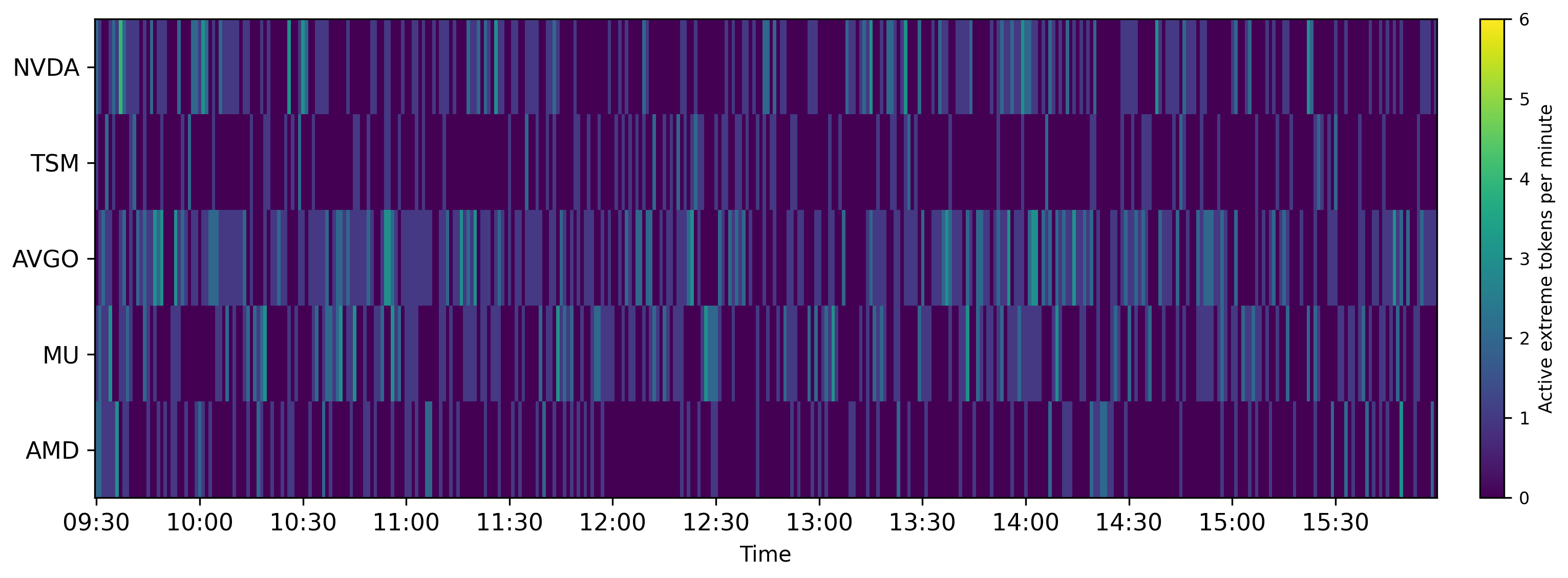}
    \caption{\footnotesize Intraday profile of dominant-exceedance tokens assigned to \textrm{NVDA}, \textrm{TSM}, \textrm{AVGO}, \textrm{MU}, and \textrm{AMD} on 23rd March 2020 (COVID-19 market crash) under the $q=0.98$ specification. Each cell reports the number of ten-second dominant-exceedance tokens aggregated within the corresponding minute.}
    \label{fig:intraday}
\end{figure}

To assess tail heaviness, we use the Hill estimator \citep{hill1975simple},
\begin{equation*}
    \widehat{\xi}_j(k)
    = \frac{1}{k}\sum_{i=1}^{k}
    \log\!\left(\frac{Y_{j,(m_j-i+1)}}{Y_{j,(m_j-k)}}\right),
\end{equation*}
where $Y_{j,(r)}$ is the $r$th smallest of the $m_j$ positive observations for stock $j$, and $1\leq k<m_j$. Across the five stocks, the marginal Hill estimates \citep{hill1975simple} range from approximately $0.26$ to $0.45$.

\begin{figure}[!htbp]
    \centering
    \includegraphics[width=0.99\linewidth]{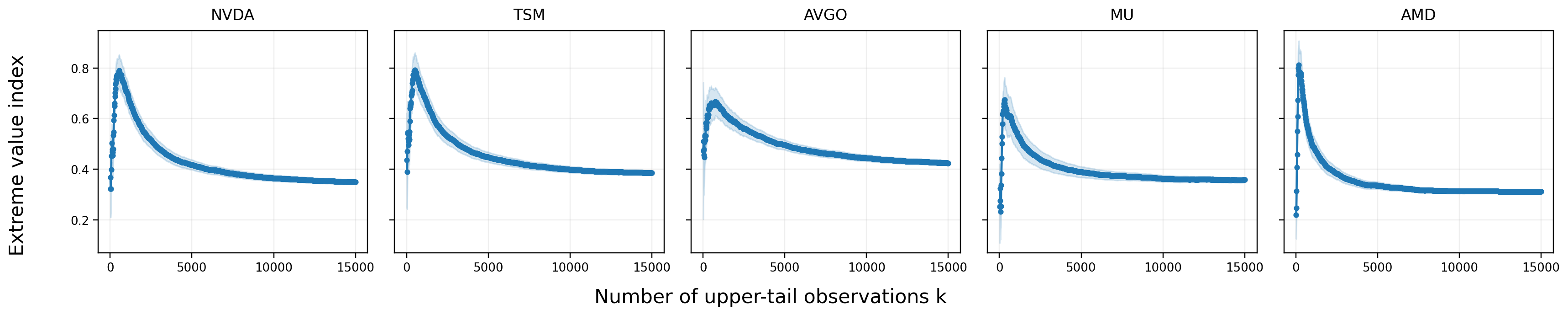}
    \caption{\footnotesize Marginal Hill estimator with its $95\%$ confidence interval for each stock. The estimator is evaluated at every $1{,}000$ values of $k$ spanning the largest $1$ to $15{,}000$ upper-tail observations out of 1.19 million.}
    \label{fig:eda}
\end{figure}

As alternatives to the empirical distribution function in \eqref{eq:empirical_cdf}, \tablename~\ref{tab:mc_fixed_empirical_robustness} compares extremal token sequences obtained from a stationary generalized Pareto model \citep[Chapter~4]{coles2001ismev} and time-varying kernel estimates $\widehat F_{t,j}$ \citep{harvey2012kernel}. The latter give greater weight to recent observations.

\begin{figure}[!htbp]
    \centering
    \includegraphics[width=0.75\linewidth]{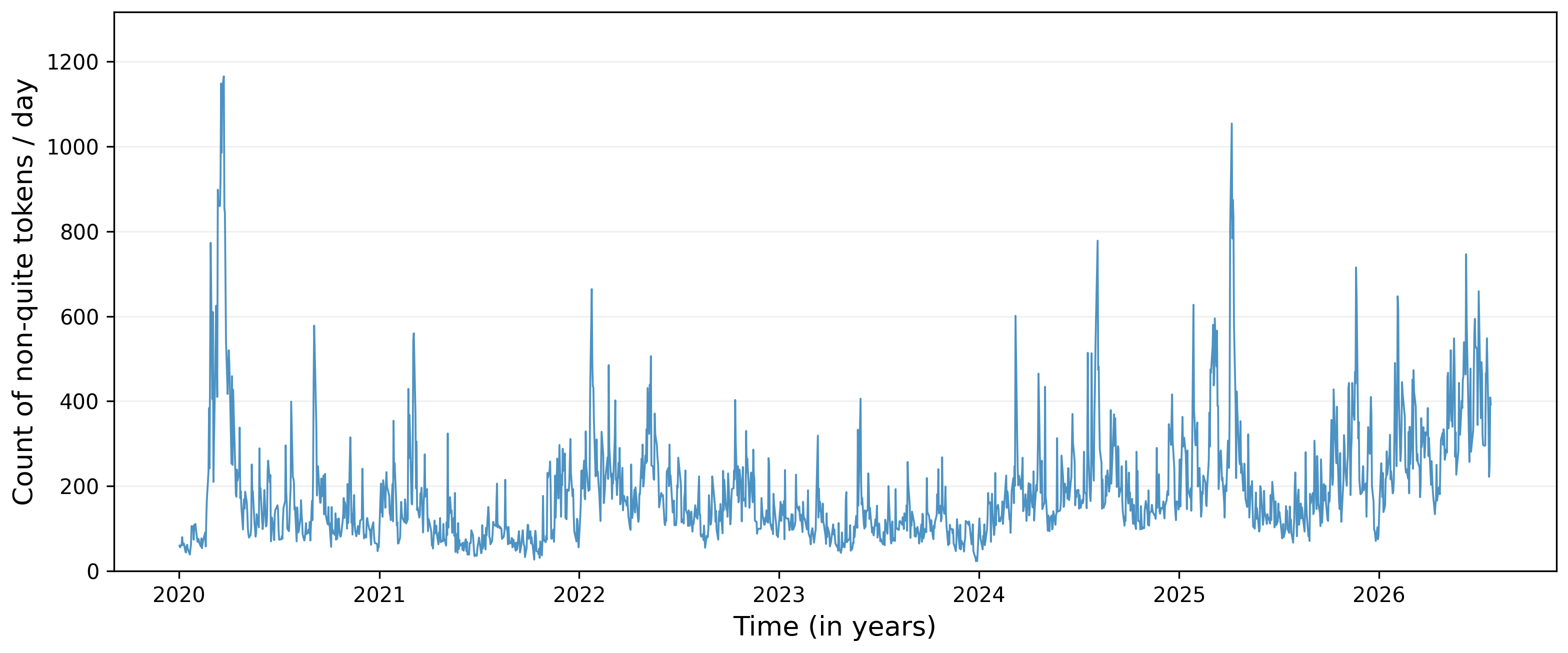}
    \caption{\footnotesize Number of non-quiet tokens per day from 2nd January 2020 to 24th July 2026 for the five selected semiconductor stocks.}
    \label{fig:timeseries}
\end{figure}

We fit a Student-$t$ VAR model to the semiconductor stocks' log returns, considering $p\in\{1,3,5,7,10\}$ and $\nu\in\{2,3,4\}$. The validation trading days were used for model selection, with each candidate evaluated by mapping its Monte Carlo one-step predictive distribution through the frozen tokenizer and computing validation next-token cross-entropy loss. Hence, the minimum validation cross-entropy was attained by the VAR($10$) model with $\nu=2$, with a validation loss of $0.554$. \tablename~\ref{tab:multi_panel_metrics} shows that the optimal-shot transformer model provides substantially better performance than the parametrized heavy-tailed VAR($10$) benchmark.

\begin{table}[!htbp]
\centering
\scriptsize
\caption{\footnotesize Comparison of transition probability metrics ($\text{d}_{\mathrm{TV}}$, $\text{d}_{\mathrm{KL}}$, and relative discrepancy, $1-\pib^{\mathrm{gen}}/\pib^{\mathrm{obs}}$) across checkpoints and the Student-$t$ VAR($10$) baseline at temperature $T=1$. Lower values indicate closer agreement with empirical transition probabilities.}\label{tab:multi_panel_metrics} \vspace{0.3cm}
\resizebox{\textwidth}{!}{
\begin{tabular}{clcccccc}
\toprule
Metric & Previous Token & Zero-shot & One-shot & Few-shot & Optimal-shot & Non-pretrained & Student-$t$ VAR \\
\midrule
\multirow{6}{*}{\makecell[l]{Rowwise Total Variation \textbf{($\text{d}_{\mathrm{TV}}$)}}}
& $\mathrm{Quiet}$ & $0.1147$ & $0.0796$ & $0.0200$ & $\mathbf{0.0252}$ & $0.1147$ & $0.1887$ \\
& $\mathrm{NVDA}$  & $0.0756$ & $0.1163$ & $0.1319$ & $\mathbf{0.0757}$ & $0.0756$ & $0.2384$ \\
& $\mathrm{TSM}$   & $0.0759$ & $0.0714$ & $0.1049$ & $\mathbf{0.0590}$ & $0.0759$ & $0.1830$ \\
& $\mathrm{AVGO}$  & $0.0701$ & $0.1127$ & $0.0806$ & $\mathbf{0.0409}$ & $0.0701$ & $0.2252$ \\
& $\mathrm{MU}$    & $0.0580$ & $0.1031$ & $0.0307$ & $\mathbf{0.0246}$ & $0.0580$ & $0.2143$ \\
& $\mathrm{AMD}$   & $0.0451$ & $0.0836$ & $0.0906$ & $\mathbf{0.0444}$ & $0.0451$ & $0.1727$ \\
\midrule
\multirow{6}{*}{\makecell[l]{Kullback--Leibler Divergence \textbf{($\text{d}_{\mathrm{KL}}$)}}} 
& $\mathrm{Quiet}$ & $0.0507$ & $0.0266$ & $0.0042$ & $\mathbf{0.0049}$ & $0.0507$ & $0.2134$ \\
& $\mathrm{NVDA}$  & $0.0682$ & $0.0928$ & $0.0837$ & $\mathbf{0.0638}$ & $0.0682$ & $0.3218$ \\
& $\mathrm{TSM}$   & $0.0431$ & $0.0327$ & $0.0448$ & $\mathbf{0.0226}$ & $0.0431$ & $0.2170$ \\
& $\mathrm{AVGO}$  & $0.0502$ & $0.0562$ & $0.0330$ & $\mathbf{0.0156}$ & $0.0502$ & $0.2568$ \\
& $\mathrm{MU}$    & $0.0170$ & $0.0365$ & $0.0043$ & $\mathbf{0.0044}$ & $0.0170$ & $0.2934$ \\
& $\mathrm{AMD}$   & $0.0129$ & $0.0339$ & $0.0222$ & $\mathbf{0.0089}$ & $0.0129$ & $0.1904$ \\
\midrule
\multirow{6}{*}{\makecell[l]{Mean rowwise relative discrepancy}} 
& $\mathrm{Quiet}$ & $0.3978$ & $0.3729$ & $\mathbf{0.1711}$ & $0.2093$ & $0.3978$ & $0.5920$ \\
& $\mathrm{NVDA}$  & $0.3456$ & $0.3602$ & $0.5340$ & $\mathbf{0.3315}$ & $0.3456$ & $0.7709$ \\
& $\mathrm{TSM}$   & $0.8848$ & $0.6553$ & $0.8785$ & $\mathbf{0.6251}$ & $0.8848$ & $0.7941$ \\
& $\mathrm{AVGO}$  & $1.2472$ & $0.8184$ & $0.9872$ & $\mathbf{0.5645}$ & $1.2472$ & $0.5465$ \\
& $\mathrm{MU}$    & $0.3981$ & $0.3985$ & $0.1810$ & $\mathbf{0.1774}$ & $0.3981$ & $0.6740$ \\
& $\mathrm{AMD}$   & $0.3858$ & $0.4457$ & $0.3851$ & $\mathbf{0.2584}$ & $0.3858$ & $0.5430$ \\
\bottomrule
\tabularnewline
\end{tabular}}
\end{table}
\begin{figure}[!htbp]
    \centering    
    \includegraphics[width=0.95\linewidth]{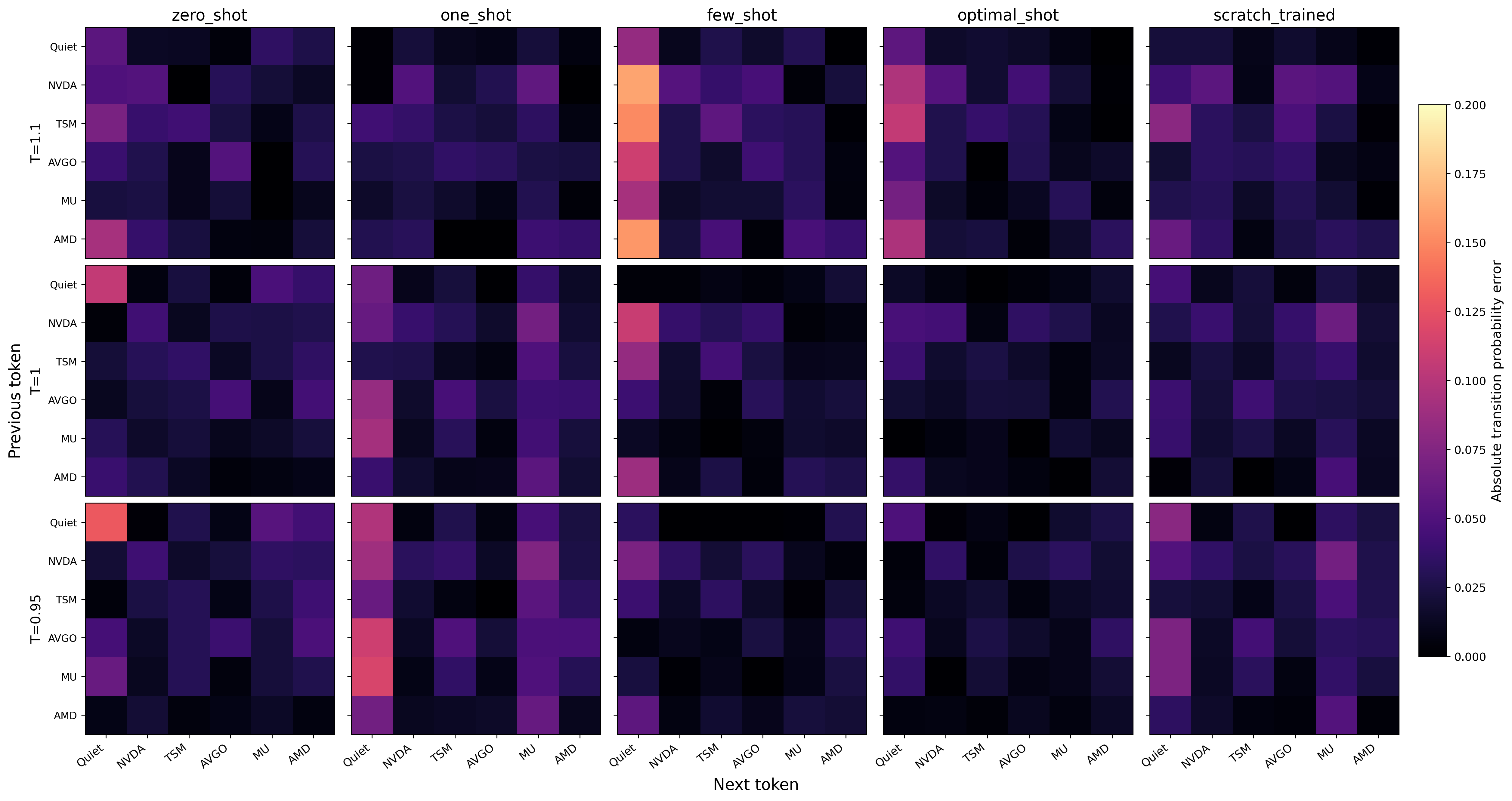}
    \caption{\footnotesize Observed and generated next-token transition structure. Each panel reports the absolute transition probability error. Rows correspond to temperatures $T\!=\!1.1$, $1$, and $0.95$, from top to bottom; columns correspond to the zero-shot, one-shot, few-shot, optimal-shot, and non-pretrained models, from left to right.}
    \label{fig:transition}
\end{figure}
\end{document}